%% file: main.tex
\documentclass{lcs2}
\input{math_commands}

\usepackage{url}
\usepackage{todonotes}
\usepackage{booktabs}
\usepackage{graphicx}
\usepackage{amsmath,amsfonts,amssymb,pifont}
\usepackage{algorithm,algpseudocode}
\usepackage{enumitem}
\usepackage{bm} 
\newcommand{\pmv}[2]{\ $#1\ {\pm\ #2}$}

\newcommand{\grade}{\texttt{GRADE}\xspace}
\newcommand{\cogrpo}{\texttt{CoGRPO}\xspace}
\newcommand{\cmark}{\ding{51}}
\newcommand{\xmark}{\ding{55}}
\usepackage{xspace,mfirstuc,tabulary}
\usepackage{graphicx}
\usepackage{subcaption}
\usepackage{booktabs} 
\usepackage{tabularx}
\usepackage{colortbl}
\usepackage{hyperref}
\usepackage{url}
\usepackage{latexsym}
\usepackage{todonotes}
\usepackage{multirow}
\usepackage{mathtools}
\usepackage{amsfonts,amsmath,amssymb,amsthm}
\usepackage{enumitem}

\usepackage{xcolor}
\definecolor{heatlow}{RGB}{245,245,245}
\definecolor{heathigh}{RGB}{33,113,181}

\usepackage{listings}
\usepackage[capitalize,noabbrev]{cleveref}

\papergithub{https://github.com/parmanu-lcs2/grade}  

\papertitle{\textit{Route}, \textit{Communicate}, and \textit{Reason}: Gated Routing and Adaptive Depth for Efficient Multi-Agent Reasoning} 
\papershorttitle{Gated Routing and Adaptive Depth for Efficient Multi-Agent Reasoning}  

\papershortauthors{S. Ghosh and T. Chakraborty} 
\paperauthors{%
  Sudipto Ghosh \textsuperscript{\tiny 1} \quad
  Tanmoy Chakraborty \textsuperscript{\tiny {1,2}}%
} 

\paperaffil{\textsuperscript{1} Yardi School of Artificial Intelligence, Indian Institute of Technology Delhi\\
\textsuperscript{2} Department of Electrical Engineering, Indian Institute of Technology Delhi} 
\paperemails{{\ttfamily sudipto.ghosh@scai.iitd.ac.in, tanchak@iitd.ac.in}} 

\paperabstract{ 
  Multi-agent ensembling multiplies active parameters and inference cost
without answering three basic questions: \emph{which} agents to consult,
\emph{how deeply} a query should traverse a hierarchy of agents, and \emph{when}
inter-agent communication is worth its cost.
We present \grade{} (\textbf{G}ated \textbf{R}outing and \textbf{A}daptive
\textbf{D}epth for \textbf{E}fficient Reasoning), a hierarchical multi-agent
system in which four lightweight learned gates jointly govern agent selection,
hierarchy depth, inter-agent communication, and branch pruning.
Training uses \texttt{CoGRPO}  (Collaborative Group-Relative Policy
Optimization), a novel critic-free recipe that adapts GRPO to multi-agent hierarchies and assigns a shared advantage signal
to every gate and agent that participated in a rollout.
Agent models are drawn from a hot-swappable Expert Registry; per-agent
calibration maps allow experts to be replaced at inference time
without retraining.
At $\sim$17B average active parameters, \grade{} outperforms all baselines on
GSM8K, MMLUPro, and GPQA, surpassing the strongest baseline by 4.8 points
on MMLUPro at half the active compute.
On AIME-2025, where model depth dominates, \grade{} remains competitive to
existing frameworks.
Ablations isolate the hierarchy and masked cross-attention as the largest
contributors to accuracy, and show that per-agent calibration is necessary
for safe hot-swapping.
} 

\begin{document}

\makelabtitle

\section{Introduction and Prior Art}
\label{sec:intro}

Each successive doubling of parameters in Large Language Models (LLMs) yields diminishing
 returns on popular benchmarks while linearly inflating memory and inference
latency~\citep{wei2022emergent}.
This cost-quality tradeoff has pushed the field toward \emph{multi-agent}
LLM coordination, where a pool of smaller, specialized models
collaborates to outperform a larger monolith at a fraction of the
active-parameter budget.

\begin{figure*}[t]
    \centering
    \includegraphics[width=\textwidth]{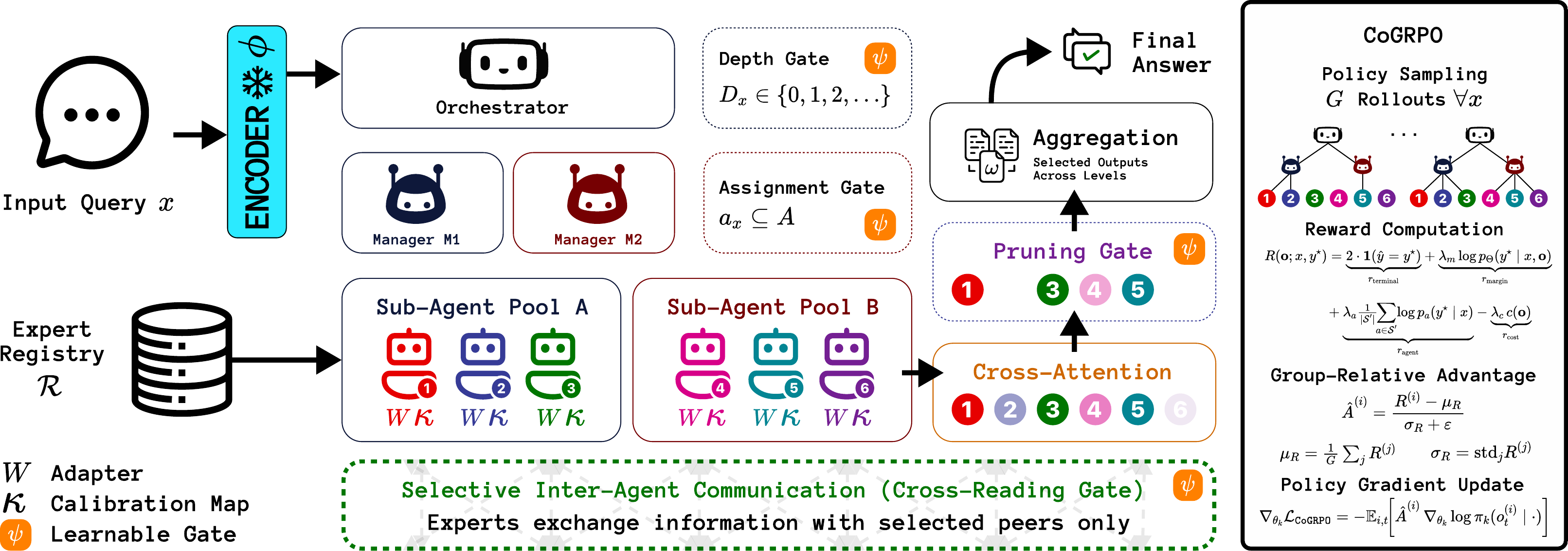}
    \caption{
\textbf{Overview of \grade.}
An encoded query passes to a lightweight \emph{orchestrator}, whose \emph{Depth
Gate} sets the reasoning depth. The orchestrator routes the query to \emph{manager} modules, whose per-sub-pool grouping signals bias an \emph{Assignment Gate} that activates only the most relevant expert sub-agents from an expert registry. Experts exchange messages selectively through the \emph{Cross-Reading Gate}, their outputs are then fused by masked cross-attention, and pass through a \emph{Pruning Gate} that drops low-utility branches before aggregation. Lightweight adapters and per-agent calibration maps allow experts to be swapped without retraining. \cogrpo{} jointly optimizes routing, depth, communication, pruning, and aggregation by sampling coordination rollouts,
computing a shared group-relative advantage, and updating all participating gates
and agents with a single policy-gradient step.
}
    \label{fig:main}
\end{figure*}

Prior multi-agent coordination strategies span several paradigms.
General-purpose agent frameworks such as AutoGen~\citep{wu2023autogen},
MetaGPT~\citep{hong2024metagpt}, CAMEL~\citep{li2023camel}, and
ChatDev~\citep{qian2024chatdev} orchestrate role-specialized LLMs through
hand-designed communication protocols, but their topology and routing are
fixed by the prompt rather than learned.
\emph{Debate and critique methods}~\citep{du2023debate,liang2023encouraging} allow multiple model instances to exchange arguments before reaching a consensus, improving factuality at the cost of requiring multiple rounds of interaction and back-and-forth communication with exploding context lengths.
\emph{Iterative self-refinement}~\citep{madaan2023selfrefine} uses extra
compute to improve a single model's output, but cannot exploit a heterogeneous
pool; reasoning-and-acting loops~\citep{yao2023react} interleave tool calls
but still operate on a single model.
\emph{Ensemble and fusion} approaches~\citep{jiang2023llmblender,wang2023selfconsistency} rank, vote over, or
fuse outputs from independently queried models; LLM-Blender~\citep{jiang2023llmblender}
adds pairwise reranking and generative fusion, but routing is fixed rather
than learned.
\emph{Mixture-of-Agents}~\citep{wang2024moa} uses layered aggregation to
refine outputs across rounds, but every agent processes every query, so
compute is not adaptive.
\emph{Evolutionary orchestration} methods such as EvoAgent~\citep{tang2024evoagent} and
Puppeteer~\citep{dang2025puppeteer}, restructure agent graphs dynamically;
Puppeteer achieves strong results via RL-trained orchestration of a flat pool
of parameter-matched models.
On the training side, MAPoRL~\citep{park2025maporl} jointly trains multiple
LLMs with multi-agent PPO, an influence-aware verification reward, and a GAE
critic, establishing that joint co-training is {necessary} -- SFT on
expert trajectories alone does not produce collaborative behavior.
\citet{liu2025magrpo} applied mean-subtracted group advantages (MAGRPO, without
std-normalization or clipping) to two-agent homogeneous systems for writing
and coding.
Both works assume flat, homogeneous pools with symmetric communication and a
fixed number of agents; neither handles hierarchical routing, selective
inter-agent communication, or heterogeneous expert management.

\begin{table*}[t]
\centering
\caption{Accuracy (\%) across four reasoning benchmarks averaged over three training seeds.
  \textbf{Act.\ Params} is the expected active-parameter count per query.
  \textbf{Overall} is the macro-average across benchmarks.
  AIME-2025 accuracy is additionally averaged over 16 decoding samples per
  problem to control small-$N$ variance (30 problems total).
  \textbf{Bold} denotes the best result per column.
  On AIME-2025, Puppeteer's larger active compute yields the highest
  accuracy; \grade{} leads on all other benchmarks.
  $^\dagger$MAGRPO \citep{liu2025magrpo} originally targets writing/coding with
  2$\times$Qwen3-1.7B; we adapt their algorithm with 2$\times$Qwen2.5-7B-Instruct
  for a fair method comparison on reasoning tasks.
  $^\ddagger$MAPoRL \citep{park2025maporl} uses 3$\times$Qwen2.5-7B-Instruct
  (original paper: Phi-3-mini) for a fair method comparison.}
\label{tab:main}
\scalebox{0.85}{
\begin{tabular}{l r cccc c}
\toprule
\textbf{Method} & \textbf{Act.\ Params} &
  \textbf{GSM8K} & \textbf{MMLUPro} & \textbf{GPQA} &
  \textbf{AIME-2025} & \textbf{Overall} \\
\midrule
Qwen2.5-7B-Instruct
  & 7B  & \pmv{85.4}{0.4} & \pmv{44.1}{0.7} & \pmv{28.6}{1.6} & \pmv{6.7}{2.3}  & \pmv{41.2}{1.3} \\
Qwen2.5-7B-Instruct + CoT
  & 7B  & \pmv{88.7}{0.5} & \pmv{50.3}{0.8} & \pmv{33.8}{1.7} & \pmv{9.1}{2.8}  & \pmv{45.5}{1.5} \\
Self-Refine
  & 7B  & \pmv{90.2}{0.4} & \pmv{56.7}{0.7} & \pmv{38.2}{1.9} & \pmv{12.4}{2.9} & \pmv{49.4}{1.5} \\
Llama-3.1-8B-Instruct
  & 8B  & \pmv{82.1}{0.6} & \pmv{41.5}{0.8} & \pmv{25.1}{1.8} & \pmv{5.4}{2.2}  & \pmv{38.5}{1.4} \\
Mixtral-8$\times$7B
  & $\sim$13B & \pmv{79.3}{0.7} & \pmv{43.0}{0.8} & \pmv{28.4}{1.8} & \pmv{5.9}{2.4} & \pmv{39.2}{1.4} \\
Ensemble-4 Majority Vote
  & $\sim$28B & \pmv{91.8}{0.4} & \pmv{62.1}{0.7} & \pmv{43.2}{1.8} & \pmv{18.3}{3.4} & \pmv{53.9}{1.6} \\
LLM-Blender
  & $\sim$28B & \pmv{92.4}{0.3} & \pmv{68.7}{0.6} & \pmv{47.6}{1.9} & \pmv{21.8}{3.6} & \pmv{57.6}{1.6} \\
EvoAgent
  & $\sim$28B & \pmv{91.9}{0.4} & \pmv{65.8}{0.7} & \pmv{45.3}{1.9} & \pmv{20.4}{3.5} & \pmv{55.9}{1.6} \\
MAGRPO$^\dagger$
  & $\sim$14B & \pmv{91.2}{0.5} & \pmv{65.2}{0.8} & \pmv{44.8}{2.0} & \pmv{18.9}{3.5} & \pmv{55.0}{1.7} \\
MAPoRL$^\ddagger$
  & $\sim$21B & \pmv{92.1}{0.4} & \pmv{71.8}{0.7} & \pmv{51.2}{2.0} & \pmv{22.1}{3.6} & \pmv{59.3}{1.7} \\
Puppeteer
  & $\sim$28B & \pmv{93.7}{0.3} & \pmv{73.4}{0.6} & \pmv{54.1}{1.9} & \pmv{\mathbf{27.2}}{3.8} & \pmv{62.1}{1.7} \\
\midrule
\textbf{\grade{} (Ours)}
  & $\sim$17B & \pmv{\mathbf{94.8}}{0.3} & \pmv{\mathbf{78.2}}{0.6} & \pmv{\mathbf{57.4}}{1.8} & \pmv{25.3}{3.6} & \pmv{\mathbf{63.9}}{1.6} \\
\bottomrule
\end{tabular}}

\end{table*}

Three limitations recur across prior systems, though none in isolation
is entirely without precedent.
(i) \textbf{Limited depth adaptivity}: Most systems route every query
through the full agent stack regardless of difficulty.
Adaptive-computation and early-exit methods adjust per-input depth for
single models~\citep{graves2016act,dehghani2019universal,elbayad2020depth,schuster2022confident,raposo2024mod},
and cost-aware cascades route easy queries to cheaper
models~\citep{chen2023frugalgpt,ong2024routellm}, but hierarchical depth gating across a
\emph{heterogeneous} multi-agent LLM pool trained end-to-end with a
single reward signal has not been studied.
(ii) \textbf{Coarse communication control}: Systems either broadcast all
agent messages or aggregate without filtering.
Learned inter-agent communication has been studied extensively in cooperative MARL literature ~\citep{sukhbaatar2016commnet,foerster2016dial,jiang2018atoc,singh2019ic3net,DasGRBPRP19},
including attention- and gating-based schemes that decide \emph{when} to
communicate, but learning a per-pair communication
mask within a jointly trained multi-agent LLM framework remains
unexplored.
(iii) \textbf{No recalibration after expert substitution}: Prior systems lack mechanisms to recalibrate gate thresholds after an expert swap, so replacing an expert silently invalidates thresholds tuned to the original model's output distribution.
Existing systems rarely combine all three mechanisms within a unified
hierarchical framework trained end-to-end. 
Our methodological  contributions are summarized below:
\begin{enumerate}[leftmargin=*, nosep]
\item \textbf{Coordination Harness}, comprising gates trained jointly via policy gradient,
      enabling per-query compute allocation, selective communication, and
      branch pruning within a single forward pass.
\item \textbf{\cogrpo}, a multi-agent adaptation  of group-relative policy
      optimization~\citep{sheng2024grpo} that requires no critic and assigns
      advantage to all gates and agents that contributed to a
      rollout.
\item \textbf{An Expert Registry} that allows individual agent experts to
      be swapped at inference time using only 64 anchor queries for
      recalibration, without touching the gates.
\end{enumerate}

\noindent Figure~\ref{fig:main} depicts an overview of the proposed framework. We ask the following questions:

\begin{enumerate}[label=\bfseries RQ\arabic*:,itemsep=0pt,wide=0pt,listparindent=0pt] 
    \item {Can a gated hierarchical multi-agent system outperform monolithic models and flat ensembles at a lower active-parameter budget? }
    \item {What is the marginal contribution of each coordination mechanism in such a system?}
    \item {Does \cogrpo's credit assignment yield better
performance than other alternatives?}
    \item {How robust is \grade{} to runtime agent substitution, and
does per-agent calibration determine the speed of accuracy recovery?}
\item {Do harder tasks recruit more agents and deeper
hierarchy levels, and what Cross-Read frequency maximizes accuracy?}
\end{enumerate}

Our empirical analyses show that \grade{} achieves \pmv{94.8}{0.3}\% on GSM8K, \pmv{78.2}{0.6}\% on
MMLUPro, and \pmv{57.4}{1.8}\% on GPQA using only $\sim$17B average
active parameters, outperforming Puppeteer by 4.8 points on
MMLUPro and 3.3 on GPQA.
On AIME-2025, Puppeteer's brute-force 4-model ensemble retains a
1.9-point advantage, confirming that raw depth of reasoning with large
monolithic models has an edge for olympiad-level mathematics.
Calibration removal causes an 18-point accuracy drop immediately after a
hot-swap and requires 20+ queries to recover, versus 0--9 with calibration.

\begin{table*}[t]
\centering
\caption{Accuracy--efficiency trade-off on MMLUPro.
  Latency is measured wall-clock per query.
  \grade{} reports a latency \emph{range} because easy queries terminate
  at depth 0 ($M_0$ only, $\sim$3s) while the hardest queries activate
  all sub-agents ($\sim$11s).
  VRAM is peak GPU memory usage at inference.
  MAGRPO runs 2 agents in 1 round; MAPoRL runs 3 agents for 3 sequential
  discussion turns, with
  all models residing in VRAM.}
\label{tab:efficiency}
\scalebox{0.88}{
\begin{tabular}{l c c c c}
\toprule
\textbf{Method} & \textbf{Act.\ Params} & \textbf{MMLUPro} &
  \textbf{Latency (s)} & \textbf{Peak VRAM (GB)} \\
\midrule
Qwen2.5-7B-Instruct
  & 7B & \pmv{44.1}{0.7} & 1.8 & 14 \\
Qwen2.5-72B-Instruct
  & 72B & \pmv{67.3}{0.6} & 31 & 144 \\
Self-Refine (Qwen2.5-7B)
  & 7B & \pmv{56.7}{0.7} & 8.7 & 14 \\
Mixtral-8$\times$7B
  & $\sim$13B & \pmv{43.0}{0.8} & 2.5 & 94 \\
Ensemble-4 Majority Vote
  & $\sim$28B & \pmv{62.1}{0.7} & 1.7 & 56 \\
LLM-Blender
  & $\sim$28B & \pmv{68.7}{0.6} & 14 & 59 \\
EvoAgent
  & $\sim$28B & \pmv{65.8}{0.7} & 38 & 56 \\
MAGRPO$^\dagger$
  & $\sim$14B & \pmv{65.2}{0.8} & 4.2 & 28 \\
MAPoRL$^\ddagger$
  & $\sim$21B & \pmv{71.8}{0.7} & 18 & 42 \\
Puppeteer
  & $\sim$28B & \pmv{73.4}{0.6} & 13 & 56 \\
\midrule
\textbf{\grade{} (Ours)}
  & $\sim$17B & \pmv{\mathbf{78.2}}{0.6} & 3.1--11.4 & \textbf{52} \\
\bottomrule
\end{tabular}}

\end{table*}
\begin{table*}[t]
\centering
\caption{Hot-swap recovery on MMLUPro (accuracy \%).
  \emph{Preswap} accuracy before substitution is 78.2\%; \emph{Acc@$n$} is
  accuracy $n$ batches after substitution; \emph{Q2R (Queries to Recover)} is
  the number of evaluation batches until accuracy returns within $0.5\%$
  of preswap.
  All calibration setups calibrate on 64 anchor queries after the swap.
 }
\label{tab:swap}
\scalebox{0.9}{
\begin{tabular}{l l l c  c c c}
\toprule
\textbf{Swap Type} & \textbf{Replaced} & \textbf{Replacement}
  & \textbf{Acc@1} & \textbf{Acc@5} &
  \textbf{Acc@10} & \textbf{Q2R} \\
\midrule
Same Model      & Qwen2.5-7B & Qwen2.5-7B      & 78.1 & 78.2 & 78.3 & 0 \\
Same Family     & Qwen2.5-7B & Qwen3-7B-Instruct & 75.4 & 77.8 & 78.1 & 4 \\
Same Size       & Qwen2.5-7B & Llama-3.1-8B   & 71.8 & 75.2 & 77.9 & 7 \\
Local$\to$API   & Qwen2.5-7B & GPT-4o-mini     & 74.7 & 77.8 & 78.0 & 3 \\
API$\to$Local   & GPT-4o-mini & Qwen2.5-7B    & 69.3 & 74.1 & 77.8 & 9 \\
No Calibration  & Any        & Any (calib.\ off)  & 60.1 & 66.8 & 71.4 & $>$20 \\
\bottomrule
\end{tabular}}

\end{table*}

\section{Methodology}
\label{sec:method}

\grade{} routes each query through a shared encoder, a lightweight orchestrator $M_0$, and a hierarchical team of managers and sub-agents. Four learned gates select active agents, routing depth, inter-agent communication, and branches to prune. The entire system is trained end-to-end with \cogrpo{} using a correctness reward.

\subsection{Architecture}
\label{sec:arch}

Let $x$ denote an input query (text prompt or numeric embedding).
A shared encoder produces a query representation $\bm{h} = \mathrm{Enc}_\phi(x) \in \mathbb{R}^{d}$.
The system is organized as a three-level hierarchy:
\begin{itemize}[leftmargin=*, nosep]
  \item \textbf{Level 0 -- Orchestrator $\mathbf{M_0}$.} A small language model
        that consumes $\bm{h}$ and emits routing decisions or a direct
        answer.
  \item \textbf{Level 1 -- Managers.} $N_M{=}2$ manager modules, each a
        trainable two-layer MLP that projects $\bm{h}$ to a manager
        representation $\bm{h}_m \in \mathbb{R}^d$. Each oversees a disjoint sub-pool of sub-agents and adds a grouping 
        signal to the Assignment Gate scoring (Section~\ref{sec:gates})  for that sub-pool.
        Managers generate no tokens -- generation is done by the frozen
        experts -- and act as a learnable grouping mechanism that
        helps multi-domain questions spanning related specialists.
  \item \textbf{Level 2 -- Sub-Agents.} Specialized expert agents,
        each backed by a (possibly distinct) frozen language model drawn
        from the Expert Registry $\mathcal{R}$. Every sub-agent $a$ carries lightweight trainable components -- an
        adapter $\Phi_a$ (frozen expert$\to$ $d$-space), an answer head
        $\bm{W}_a$, and a calibration map $\kappa_a$ (Section \ref{sec:calib}).
        The frozen expert of agent $a$ is $\{\theta_a\}$;
        agents that resolve to the same model identifier share a \emph{single
        in-memory copy}, so VRAM is proportional to
        the number of \emph{distinct} models, not the number of agents.
        We write $\mathcal{A} = \{a_1,\ldots,a_N\}$ and denote agent $a$'s
        learned registry descriptor embedding $\bm{e}_a \in \mathbb{R}^d$.
\end{itemize}

\subsection{Coordination Gates}
\label{sec:gates}

All gates are small two-layer MLPs with GELU activations, parameterized
by $\psi = \{\psi^A, \psi^D, \psi^R, \psi^P\}$, jointly optimized with
the rest of the system via \texttt{CoGRPO}.

\begin{algorithm}[t]
\caption{\cogrpo{} Training Step}
\label{alg:cogrpo}
\small
\begin{algorithmic}[1]
\Require Query $x$, gold $y^\star$, group size $G$, clip $\eta$,
         KL weight $\beta$, dense-reward weights
         $(\lambda_m,\lambda_a,\lambda_c)$, inner epochs $E$,
         frozen registry $\mathcal{R}$
\Ensure Updated parameters $\Theta = \{\phi,\psi,\omega\}$
\State $h \gets \mathrm{Enc}_\phi(x)$ \Comment{shared encoding}
\For{$i = 1,\ldots,G$} \Comment{sample group of rollouts}
  \State Sample joint action $\bm{o}^{(i)}\sim\pi_\Theta(\cdot\mid x)$
         via Eq.~\eqref{eq:joint_policy}
  \State Run agents in $\mathcal{S}'^{(i)}$;\;
         $\hat{y}^{(i)}\gets\mathrm{Combine}_\omega(\ldots)$
  \State Compute dense reward $R^{(i)}$ via Eq.~\eqref{eq:reward_formal}
  \State Store $\mathrm{lp}^{(i)} = \log\pi_\Theta(\bm{o}^{(i)}\mid x)$
\EndFor
\State $\hat{A}^{(i)} \gets (R^{(i)}-\mu_R)/(\sigma_R+\varepsilon)\;\;\forall i$
       \Comment{advantage, Eq.~\eqref{eq:group_stats}}
\For{$e = 1,\ldots,E$} \Comment{inner PPO epochs}
  \For{$i = 1,\ldots,G$}
    \State $\varrho_t^{(i)} \gets \exp(\log\pi_\Theta(o_t^{(i)}) - \mathrm{lp}_t^{(i)})$
    \State Accumulate $\mathcal{L}_\text{CoGRPO}$ (Eq.~\eqref{eq:cogrpo_loss})
           and gate regularizers
  \EndFor
  \State $\Theta \gets \Theta - \alpha\nabla_\Theta\mathcal{L}(\Theta)$
         via Eq.~\eqref{eq:total_loss}
\EndFor
\end{algorithmic}
\end{algorithm}

\paragraph{(i) Assignment Gate.}
The Assignment Gate selects which sub-agents are activated for query $x$.
It scores each agent $a$ by the affinity between the query and the agent
descriptor, plus the grouping signal contributed by the manager $m(a)$
that oversees $a$'s sub-pool:
\begin{equation}
s_a = \frac{\langle \bm{W}^A\bm{h}, \bm{e}_a\rangle}{\sqrt{d}}
      + \frac{\langle \bm{h}_{m(a)}, \bm{e}_a\rangle}{\sqrt{d}},
\label{eq:assign}
\end{equation}
where $m(a)\in\{1,\dots,N_M\}$ is the unique manager whose sub-pool
contains agent $a$, so $\bm{h}_{m(a)}$ is the same manager representation
(Section~\ref{sec:arch}) (every
agent in a given sub-pool therefore shares one $\bm{h}_{m(a)}$). The
manager term biases the scores of a whole sub-pool together, letting a
manager recruit its related specialists jointly on multi-domain queries. 
In \emph{dynamic-topology} mode, each agent is independently
sampled from $\text{Bernoulli}(\sigma(s_a))$, with at least one agent
guaranteed; a fixed top-$k$ mode is also supported for the ablation
in Table~\ref{tab:ablation_rl}.
Load-balance regularization prevents collapse onto a small favorite set:
\begin{equation}
\mathcal{L}_{\text{bal}}
= \lambda_{\mathrm{bal}} \cdot N \sum_{a} f_a p_a,
\label{eq:balance}
\end{equation}
where $f_a$ is the empirical routing fraction and $p_a$ is the mean
gate probability for agent $a$ across the batch.

\paragraph{(ii) Depth Gate.}
The Depth Gate controls how many hierarchy levels a query traverses,
implementing adaptive computation analogous to Mixture of
Depths~\citep{raposo2024mod} and confident early
exit~\citep{schuster2022confident}:
\begin{equation}
\begin{gathered}
\bm{g}^D = \mathrm{softmax}\big(\mathrm{MLP}_{\psi^D}(\bm{h})\big)
\in \Delta^{D-1}, \quad
\ell^\star = \arg\max_{\ell\in\{0,\dots,D-1\}} g^D_\ell,
\end{gathered}
\end{equation}
where $D$ is the number of hierarchy depth levels and $\Delta^{D-1}$
denotes the corresponding probability simplex (the set of categorical
distributions over the $D$ levels). In our instantiation $D{=}3$ -- Depth $\ell^\star{=}0$ terminates at $M_0$ (cheapest path);
$\ell^\star{=}1$ involves managers; $\ell^\star{=}2$ involves the full
sub-agent pool.
At training time, $\ell^\star$ is sampled for exploration.
A depth penalty $\mathcal{L}_{\mathrm{depth}} = \lambda_D\mathbb{E}[\ell^\star]$
discourages unnecessary descent. 

\paragraph{(iii) Cross-Read Gate.}
Once sub-agents produce message embeddings $\bm{m}_a \in \mathbb{R}^d$,
the Cross-Read Gate determines which pairs of agents may exchange
information before the combination.
For each ordered pair $(a,b)$, it emits a binary read decision via a
Gumbel-sigmoid relaxation~\citep{jang2017gumbel}:
\begin{equation}
\begin{gathered}
r_{ab} = \sigma\Big(\tfrac{1}{\nu}\big(
  \mathrm{MLP}_{\psi^R}([\bm{m}_a;\bm{m}_b]) + \epsilon\big)\Big),\\
  \epsilon \sim \mathrm{Logistic}(0,1),
\end{gathered}
\end{equation}
where $\nu{=}0.5$ is a temperature.
Agent $a$ may attend to $b$ in the cross-attention step iff $r_{ab}{>}0.5$.
A sparsity penalty $\mathcal{L}_{\mathrm{read}} = \lambda_R\sum_{a,b} r_{ab}$
keeps communication economical.

\paragraph{(iv) Prune Gate.}
Before final combination, the Prune Gate assigns each active branch a
scalar value estimate:
\begin{equation}
v_a = \sigma\big(\mathrm{MLP}_{\psi^P}([\bm{h};\bm{m}_a])\big) \in [0,1],
\end{equation}
and drops branches with $v_a < \tau^P{=}0.35$, ensuring at least one
survives.
Pruning removes uninformative branches before the expensive cross-attention
combination step, recovering most of the latency advantage in
Table~\ref{tab:efficiency}.

\subsection{Answer Combination}
\label{sec:combine}

Let $\mathcal{S}'(x)$ denote the set of surviving (assigned and un-pruned)
sub-agents.
Their calibrated messages $\tilde{\bm{m}}_a$ are stacked into
$\bm{M} \in \mathbb{R}^{|\mathcal{S}'|\times d}$ and fused by a masked
multi-head cross-attention module:
\begin{equation}
\bm{Z} = \mathrm{softmax}\left(
  \frac{(\bm{M}\bm{W}_Q)(\bm{M}\bm{W}_K)^\top}{\sqrt{d}}
  \odot \bm{R}\right)(\bm{M}\bm{W}_V),
\end{equation}
where $\bm{R}_{ab} = r_{ab}$ is the Cross-Read mask.
The pooled combined representation $\bm{z} = \mathrm{Pool}(\bm{Z})$ is
decoded to the final answer by a combination head:
\begin{equation}
\hat{y} = \mathrm{Combine}_\omega(\bm{z}, \bm{h}, \bm{m}),
\end{equation}
where $\bm{m}$ is a persistent memory buffer updated by an
exponential moving average of $\bm{z}$ within each training batch.
Concretely, $\bm{m} \leftarrow (1-\alpha)\bm{m} + \alpha\bm{z}$
is applied after every forward pass in the batch; at evaluation time
$\bm{m}$ is fixed to its end-of-training value and never updated.
Therefore, $\bm{m}$ is a batch-level running statistic and is
\emph{not} conditioned on any individual query's identity, so it
cannot cause cross-example leakage during evaluation on single-turn
benchmarks. Its benefit is a stable low-variance context bias for the
combination head rather than genuine episodic memory.

\subsection{Training with \cogrpo}
\label{sec:training}

Because all gate decisions are discrete, standard backpropagation does not
apply. Reinforcement learning has become the standard way for optimizing
non-differentiable objectives in LLMs~\citep{ouyang2022rlhf,schulman2017ppo}. We train with \cogrpo, which extends group-relative policy
optimization~\citep{sheng2024grpo,guo2025deepseekr1} to the multi-agent
setting: no critic is needed, and every gate and agent that participated in
a rollout receives the same scalar advantage.

\paragraph{Joint Policy.}
\grade{} has trainable parameters $\Theta = \{\phi, \psi, \omega\}$, where
$\phi$ are the shared encoder layers,
$\psi = \{\psi^A,\psi^D,$ $\psi^R,\psi^P\}$ are the four gate parameter sets,
and $\omega$ are the combination-head parameters; the agent adapters and
answer heads $\{(\Phi_a,\bm{W}_a)\}$ are also in $\Theta$, while all
experts $\{\theta_a\}$ remain frozen.
A single rollout for query $x$ is a sequence of stochastic decisions
\begin{equation}
\begin{gathered}
\ell^\star \sim \pi_{\psi^D}(\cdot\mid h),\quad
\mathcal{S} \sim \pi_{\psi^A}(\cdot\mid h),\quad
\bm{r} \sim \pi_{\psi^R}(\cdot\mid \bm{M}),\quad
\mathcal{K} \sim \pi_{\psi^P}(\cdot\mid h,\bm{M}),
\end{gathered}
\end{equation}
where $h = \mathrm{Enc}_\phi(x)$ and $\bm{M}\in\mathbb{R}^{N\times d}$ stacks
the agent messages. Together they define a \emph{joint policy} over all
coordination decisions in one pass:
\begin{equation}\begin{aligned}
\pi_\Theta(\bm{o}\mid x)
= &\, \pi_{\psi^D}(\ell^\star\mid h)\,
  \pi_{\psi^A}(\mathcal{S}\mid h) 
  \, \pi_{\psi^R}(\bm{r}\mid\bm{M})\,
  \pi_{\psi^P}(\mathcal{K}\mid h,\bm{M}),
\label{eq:joint_policy}
\end{aligned}
\end{equation}
where $\bm{o} = (\ell^\star,\mathcal{S},\bm{r},\mathcal{K})$ is the full
joint action and the surviving agent set is
$\mathcal{S}'(x) = \mathcal{S}\cap\mathcal{K}$.

\paragraph{Dense Reward.}
For each query $x$, we sample $G{=}8$ joint rollouts.
Each rollout $\bm{o}^{(i)}$ yields an answer $\hat{y}^{(i)}$, a compute
cost $c^{(i)}$, and per-agent answer distributions over the surviving
agents $\mathcal{S}'^{(i)}$.
The reward has four terms:
\begin{equation}
\begin{aligned}
R(\bm{o};x,y^\star)
&= \underbrace{2\cdot\mathbf{1}(\hat{y}=y^\star)}_{r_\text{terminal}}
 + \underbrace{\lambda_m \log p_\Theta(y^\star\mid x,\bm{o})}_{r_\text{margin}}\\
&\;+ \underbrace{\lambda_a \tfrac{1}{|\mathcal{S}'|}\!\!\sum_{a\in\mathcal{S}'}\!\!
    \log p_a(y^\star\mid x)}_{r_\text{agent}}
 - \underbrace{\lambda_c\, c(\bm{o})}_{r_\text{cost}},
\end{aligned}
\label{eq:reward_formal}
\end{equation}
where $c(\bm{o}) = \text{token\_cost}(\bm{o})/1000$ is the normalized cost
and all log-probabilities are clamped to $[-5,0]$ to prevent exploding
gradients when probabilities are near zero.
Scaling  $r_\text{terminal}$  to 2 ensures correctness dominates the cost
penalty for all query budgets: with $\lambda_c{=}0.05$ and a normalized cost
$c^{(i)}$ that stays in $[0.1,\,2.5]$ across all task families, the cost term
$\lambda_c c^{(i)}$ deducts at most ${\sim}0.12$ -- far smaller than the
2-point unit gap between a
correct and an incorrect terminal reward.
Without this headroom, early policies that accidentally recruit fewer agents
can become locally optimal purely from cost savings, collapsing the Depth
and Assignment Gates before the correctness signal stabilizes.
Unlike $r_\text{terminal}$, the margin term $r_\text{margin}$ is non-zero on
\emph{every} rollout, supplying a smooth gradient even when the prediction is
wrong; $r_\text{agent}$ gives each surviving agent's answer head an
independent alignment signal, penalizing free-riding where an agent passes
noisy messages and lets the combination head compensate.

\begin{table*}[t]
\centering
\caption{RL objective comparison (top) and fixed-$k$ assignment ablation (bottom).
  All objectives in the top block are applied to the same \grade{} architecture
  and agent pool; only the training algorithm changes.
  $^\dagger$~\citet{liu2025magrpo}'s group-relative advantage
  with mean-only normalization (no std-norm, no clip, no KL), ported to \grade{}'s
  joint gate policy.
  $^\ddagger$\citet{park2025maporl}'s GAE critic
  and influence-aware reward adapted to \grade{}'s hierarchical action space;
  per-agent influence approximated via cross-read frequency differential.}
\label{tab:ablation_rl}
\scalebox{0.88}{
\begin{tabular}{l ccccc}
\toprule
\textbf{Configuration} & \textbf{GSM8K} & \textbf{MMLUPro} &
  \textbf{GPQA} & \textbf{AIME-2025} & \textbf{Overall} \\
\midrule
\textbf{\cogrpo}          & \pmv{\mathbf{94.8}}{0.3} & \pmv{\mathbf{78.2}}{0.6} & \pmv{\mathbf{57.4}}{1.8} & \pmv{\mathbf{25.3}}{3.6} & \pmv{\mathbf{63.9}}{1.6} \\
MAGRPO-obj$^\dagger$ & \pmv{93.9}{0.4} & \pmv{76.0}{0.6} & \pmv{55.3}{1.9} & \pmv{23.8}{3.6} & \pmv{62.3}{1.6} \\
Shared Reward        & \pmv{93.8}{0.4} & \pmv{75.1}{0.7} & \pmv{54.2}{1.9} & \pmv{23.2}{3.7} & \pmv{61.6}{1.7} \\
 MAPoRL-obj$^\ddagger$ & \pmv{93.5}{0.5} & \pmv{73.2}{0.8} & \pmv{52.8}{2.0} & \pmv{22.9}{3.7} & \pmv{60.6}{1.8} \\
 MAPPO                & \pmv{93.1}{0.5} & \pmv{72.6}{0.8} & \pmv{51.8}{2.1} & \pmv{22.4}{3.7} & \pmv{60.0}{1.8} \\
REINFORCE            & \pmv{92.4}{0.6} & \pmv{69.8}{0.9} & \pmv{48.3}{2.2} & \pmv{20.1}{3.9} & \pmv{57.7}{1.9} \\
\midrule
$k=1$ (fixed)     & \pmv{90.2}{0.5} & \pmv{64.3}{0.9} & \pmv{42.1}{2.1} & \pmv{14.7}{3.5} & \pmv{52.8}{1.8} \\
$k=2$ (fixed)     & \pmv{93.1}{0.4} & \pmv{72.8}{0.7} & \pmv{51.4}{2.0} & \pmv{21.4}{3.6} & \pmv{59.7}{1.7} \\
$k=3$ (fixed)     & \pmv{94.3}{0.3} & \pmv{77.8}{0.6} & \pmv{56.9}{1.8} & \pmv{24.8}{3.6} & \pmv{63.5}{1.6} \\
$k=4$ (fixed)     & \pmv{94.3}{0.3} & \pmv{77.1}{0.6} & \pmv{56.2}{1.8} & \pmv{24.3}{3.5} & \pmv{63.0}{1.6} \\
$k=5$ (fixed)     & \pmv{93.8}{0.4} & \pmv{76.3}{0.6} & \pmv{55.1}{1.9} & \pmv{23.2}{3.6} & \pmv{62.1}{1.6} \\
\bottomrule
\end{tabular}}
\vspace{-3mm}
\end{table*}

\paragraph{Group-Relative Advantage.}
Instead of a learned critic, \cogrpo{} normalizes rewards within each group of $G$ rollouts:
\begin{equation}
\begin{gathered}
\hat{A}^{(i)} = \frac{R^{(i)} - \mu_R}{\sigma_R + \varepsilon},
\end{gathered}
\label{eq:group_stats}
\end{equation}
with $\mu_R = \tfrac{1}{G}\textstyle\sum_j R^{(j)}$, $
\sigma_R = \mathrm{std}_j R^{(j)}$ and $\varepsilon{=}10^{-4}$.
Every gate and agent that contributed to rollout $i$ receives
$\hat{A}^{(i)}$, which we call \emph{collaborative credit assignment}. 
When all $G$ rollouts receive the same reward, $\sigma_R{=}0$ and
$\hat{A}^{(i)}{=}0$, producing no update -- an indistinguishable
group should drive no policy change.
Mean-only normalization~\citep{liu2025magrpo} also vanishes in this case,
but its gradient magnitude scales with the absolute reward rather than the
within-group contrast; standard normalization decouples gradient magnitude
from reward scale, improving stability across tasks.

\paragraph{Relation to GRPO.}
\cogrpo{} differs from vanilla GRPO~\citep{sheng2024grpo} in two ways
specific to the multi-agent setting.
GRPO optimizes a single sequential policy; \cogrpo{} optimizes a
\emph{joint} policy $\pi_\Theta$ factored across gate distributions
(Eq.~\eqref{eq:joint_policy}), so the importance ratio $\varrho_t^{(i)}$
is decomposed per gate component rather than per output token.
GRPO's group varies only in the output sequence; our group varies
in depth, agent assignment, cross-read structure, and pruning decisions.
A naive application of GRPO to the joint policy would treat the entire
forward pass as one action, providing no gradient signal to individual
gate decisions; the per-component decomposition below distributes the
advantage to each participant.

\paragraph{Collaborative Credit Assignment.}
Counterfactual per-gate credit -- holding each gate at a do-not-participate
baseline and rerunning the pass -- would cost $O(2^4)$ forward passes per
group. Instead, the group baseline $\mu_R$ absorbs the mean contribution of
all components at once: every participant of rollout $i$ -- each gate that
made a non-trivial decision and each surviving agent $a\in\mathcal{S}'^{(i)}$
-- shares the same advantage $\hat{A}^{(i)}$, so the gradient for any
participant $\pi_k$ is
\begin{equation}
\begin{aligned}
\nabla_{\theta_k}\mathcal{L}_{\cogrpo}
= -\mathbb{E}_{i,t}\!\left[
    \hat{A}^{(i)}\,\nabla_{\theta_k}\log\pi_k(o^{(i)}_t\mid\cdot)\right],
\label{eq:coop_grad}
\end{aligned}
\end{equation}
as in team-reward cooperative MARL~\citep{yu2022mappo}, but with the
group-relative baseline in place of a centralized critic. This credit is
\emph{noisy but unbiased}: a neutral decision (e.g.\ a cross-read mask that
neither helps nor hurts) gets the same $\hat{A}^{(i)}$ as a decisive one, and the noise averages out over many (query, group) pairs.
Diversity across rollouts is preserved by the auxiliary regularizers -- the
load-balance loss (Eq.~\eqref{eq:balance}) prevents routing collapse, the
depth penalty prevents all-depth-2 solutions, and the read-sparsity penalty
prevents the Cross-Read Gate from setting all $r_{ab}{=}1$ --- so no gate
distribution degenerates: in practice $\rho\ {\approx}\ 0.5$, mean depth
${\approx}1.9$, and mean assignment $k\ {\approx}\ 2-4$ throughout training.

\paragraph{Clipped Surrogate Objective.}
With per-decision importance ratio
$\varrho_t^{(i)} = \pi_\Theta(o_t^{(i)})/\pi_{\Theta_\mathrm{old}}(o_t^{(i)})$
computed against the snapshot $\Theta_\mathrm{old}$ that generated the
rollouts:
\begin{equation}
\mathcal{L}_{\mathrm{\cogrpo}}
= -\mathbb{E}_{i,t}\Big[
    \min\big(\varrho_t^{(i)}\hat{A}^{(i)},
    \mathrm{clip}(\varrho_t^{(i)}, 1{-}\eta, 1{+}\eta)\hat{A}^{(i)}\big)
  \Big]
  + \beta\mathbb{D}_{KL}\big(\pi_\Theta\parallel\pi_{\mathrm{ref}}\big),
\label{eq:cogrpo_loss}
\end{equation}
where $\eta{=}0.2$ is the clip radius and $\beta{=}0.02$ weights the KL
divergence to the frozen reference policy $\pi_\mathrm{ref}$.
Clipping prevents any single rollout from making an excessively large
update, while the KL term prevents catastrophic forgetting of the
inductive biases of the pre-trained models.
The full training loss adds the gate regularizers and the supervised
auxiliary term:
\begin{equation}
\begin{aligned}
\mathcal{L} = \mathcal{L}_{\mathrm{\cogrpo}}
            + \mathcal{L}_{\mathrm{bal}}
            + \mathcal{L}_{\mathrm{depth}}
            + \mathcal{L}_{\mathrm{read}}
            + \tfrac{1}{2}\mathcal{L}_{\mathrm{sup}}.
\end{aligned}
\label{eq:total_loss}
\end{equation}

\begin{figure}[t]
  \centering
  \includegraphics[width=.5\textwidth]{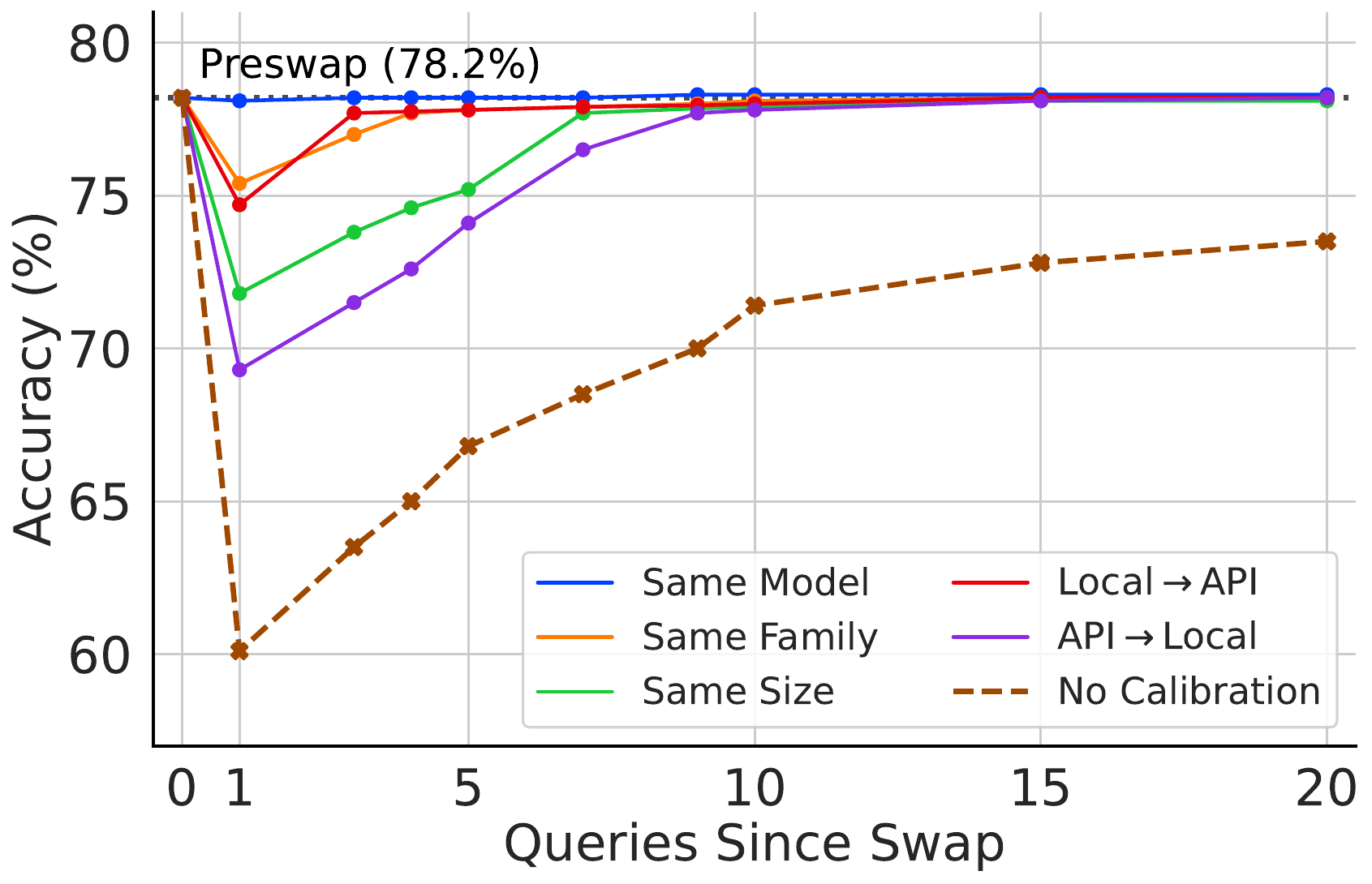}
  \caption{Accuracy recovery curves after each swap type.
    MMLUPro accuracy (\%) vs.\ evaluation batches since swap.
    The ``No Calibration'' curve remains substantially below preswap
    accuracy, while calibrated swaps converge within 9 batches.}
  \label{fig:swap}
\end{figure}

\paragraph{Supervised Auxiliary Loss.}
The differentiable components (adapters, answer heads, combination head)
also receive a cross-entropy loss $\mathcal{L}_\text{sup}$ on the
realized answer, applied only to the combination head's logits.
This stabilizes early training before \cogrpo{} can provide a useful signal.  $\mathcal{L}_\text{sup}$ gradients are \emph{not} propagated
through the gate sampling operations: gates receive gradient only from
the policy gradient term.
The dense reward components $r_\text{margin}$ and $r_\text{agent}$
overlap in intent with $\mathcal{L}_\text{sup}$ -- both encourage
high $p(y^\star)$ -- but their influence on gate decisions flows through
the advantage signal rather than through direct backpropagation, so they
are not redundant.
Sub-agent experts remain frozen throughout; only gates $\psi$, adapters,
answer heads, combination head $\omega$, and the last two encoder layers
$\phi$ are updated.

Table~\ref{tab:algo_compare} summarizes how \cogrpo{} differs from related
objectives in critic use, standard-normalization, importance-ratio clipping,
KL regularization, and multi-agent structure.

\subsection{Calibration for Hot-Swappable Experts}
\label{sec:calib}

Sub-agents are drawn from a heterogeneous registry and may be replaced at
inference time.
A 3B local model and a GPT-4o-mini API model produce embeddings that differ
in scale, variance, and alignment; substituting one for the other invalidates
the gate thresholds that were fitted during training.
Modern neural networks, including pre-trained transformers, are well known
to be miscalibrated and sensitive to such distribution
shift~\citep{guo2017calibration,desai2020calibration}, which a fixed
threshold cannot absorb.
We address this with a per-agent \emph{calibration map} $\kappa_a$, a
learnable affine normalization that brings each backend's messages into the
shared embedding space:
\begin{equation}
\tilde{\bm{m}}_a = \kappa_a(\bm{m}_a)
= \bm{\gamma}_a \odot \frac{\bm{m}_a - \hat{\bm{\mu}}_a}{\hat{\bm{\sigma}}_a} + \bm{\beta}_a,
\end{equation}
where $(\hat{\bm{\mu}}_a, \hat{\bm{\sigma}}_a)$ are running statistics
of agent $a$'s message embeddings updated with momentum $\alpha{=}0.05$,
and $(\bm{\gamma}_a, \bm{\beta}_a)$ are learned affine parameters.
After a swap, only the replaced agent's $\kappa_a$ is re-estimated using
64 anchor queries; everything else stays fixed.
Table~\ref{tab:swap} shows that removing calibration causes an immediate
18-point accuracy drop and requires more than 20 queries to recover.

\begin{figure}[t]
  \centering
  \includegraphics[width=.5\columnwidth]{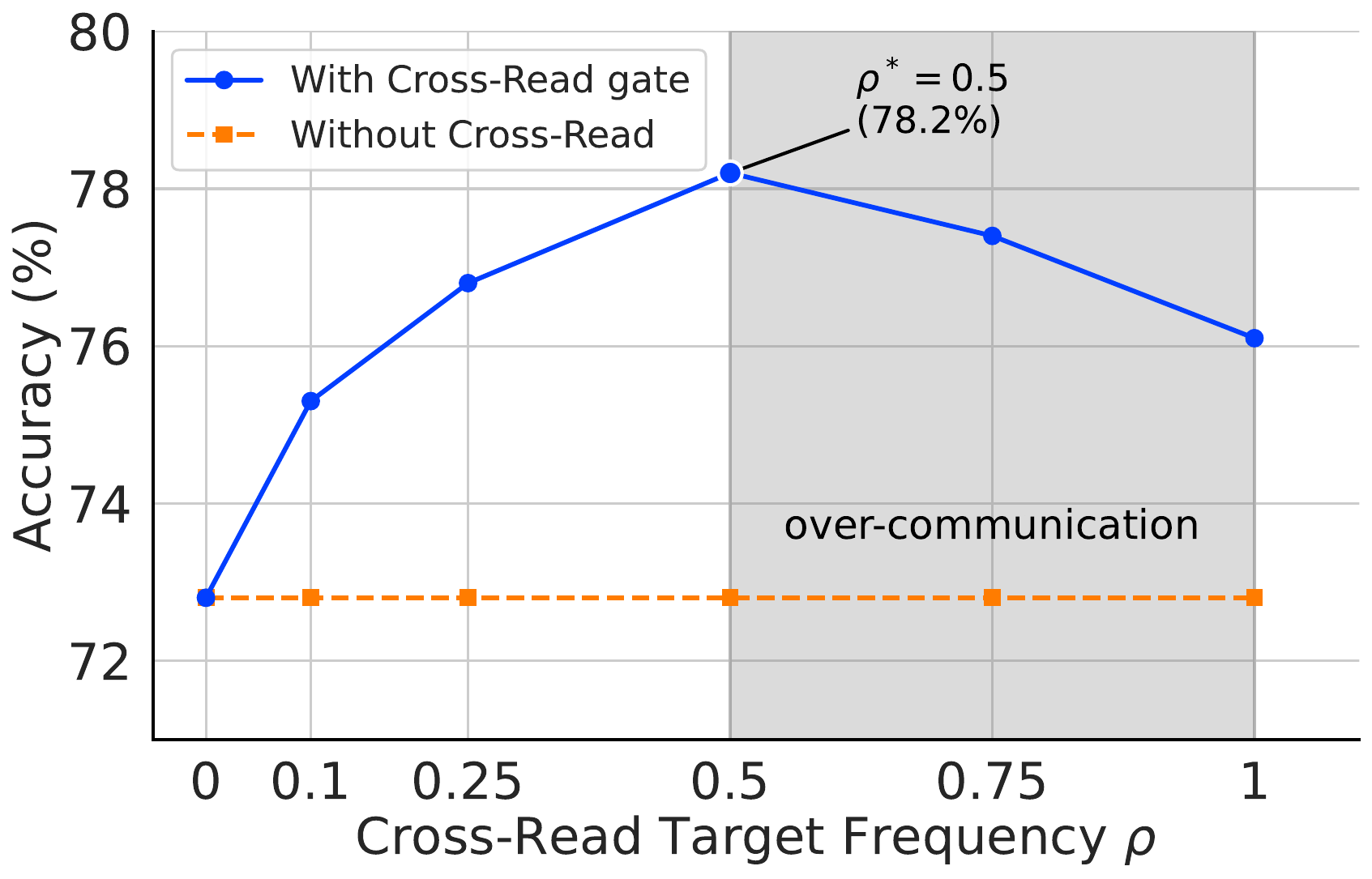}
  \caption{Cross-Read frequency vs.\ MMLUPro accuracy.
 The Cross-Read frequency $\rho$ is the fraction of permitted agent pairs.
  $\rho{=}0.0$ is equivalent to no communication; $\rho{=}1.0$ introduces
  noise from irrelevant cross-agent messages.}
  \label{fig:crossread}
\end{figure}
\section{Experiments}
\label{sec:experiments}

\subsection{Experimental Setup}

\paragraph{Datasets.}
We evaluate on four benchmarks of increasing difficulty --- {GSM8K}~\citep{cobbe2021gsm8k}, {MMLUPro}~\citep{wang2024mmlupro}, {GPQA}~\citep{rein2023gpqa} and {AIME-2025}~\citep{dekoninck2026beyond}.
All benchmarks are evaluated with zero-shot prompting unless noted;
CoT baselines use 8-shot exemplars following~\citet{wei2022cot}. Full prompt templates are listed in Appendix~\ref{app:prompts}.

\paragraph{Agent Pool.}
The \grade{} pool for our experiment comprises six sub-agents backed by three distinct frozen
models: A1--A3 share \textbf{Qwen2.5-7B-Instruct}; B1--B2 share
\textbf{Phi-3-mini-4K-Instruct}~\citep{microsoft2024phi3} (3.8B); and B3 uses
\textbf{Llama-3.2-3B-Instruct}~\citep{meta2024llama3}.
The orchestrator $M_0$ is \textbf{Qwen2.5-0.5B-Instruct}; the two managers
are lightweight trainable MLP grouping modules (Section~\ref{sec:arch}) with
no language-model backend of their own.
We define \emph{active parameters} per query as the expected total parameters
evaluated across all forward passes in a rollout ($M_0$ and the sub-agents are
language models; the managers and gates are parameter-light).
This ranges from 0.5B for a depth-0 query ($M_0$ only) to $\sim$32B when all
six sub-agents run at depth~2, and averages $\sim$17B per query.

\subsection{Token Usage Over Training}
\label{sec:e6}

\paragraph{Baselines.}
We compare against: a single Qwen2.5-7B-Instruct; Qwen2.5-7B with 8-shot
chain-of-thought~\citep{wei2022cot}; Self-Refine~\citep{madaan2023selfrefine}
with three refinement rounds; Llama-3.1-8B-Instruct~\citep{meta2024llama3};
Mixtral-8$\times$7B~\citep{jiang2024mixtral}; an ensemble of four
Qwen2.5-7B instances with majority voting; LLM-Blender~\citep{jiang2023llmblender}
with four agents; EvoAgent~\citep{tang2024evoagent} with four agents; and
Puppeteer~\citep{dang2025puppeteer} with four Qwen2.5-7B-Instruct agents
(the strongest current system).

\paragraph{Training.}
We train \grade{} for 200 gradient steps with Adam (LR~$= 10^{-4}$),
group size $G{=}8$, clip $\eta{=}0.2$, KL weight $\beta{=}0.02$, and
cost weight $\lambda_c{=}0.05$.
All experiments are run on a single NVIDIA H200 NVL 141GB GPU.
We report mean and standard deviation across 3 random seeds.
Full hyperparameters are in Appendix~\ref{app:hparams}.

\subsection{Comparison with Baselines}
\label{sec:e1}

\grade{} achieves the best overall score (\pmv{63.9}{1.6}\%) versus
Puppeteer's \pmv{62.1}{1.7}\%, at just over half the active parameters
(Table~\ref{tab:main}).

The gains over Puppeteer are largest on MMLUPro ($+4.8$ points) and GPQA
($+3.3$ points).
On MMLUPro, the Depth Gate skips sub-agent computation on easy queries,
and Cross-Read enables targeted sharing on multi-domain questions.
On GPQA, the manager layer groups domain-specialist agents before
combination, which helps boundary questions that span biology, chemistry,
and physics.
The GSM8K gap is modest ($+1.1$ points): elementary arithmetic is already
near-saturated by a single 7B model.

On AIME-2025, Puppeteer's four 8B agents retain a 1.9-point advantage
(27.2 vs.\ 25.3).
Our sub-agents (7B, 3.8B, 3B) are individually weaker at the long,
sequential reasoning chains these problems demand, and routing benefits
do not compensate for lower per-agent capacity.
Adding math-specialist backends (e.g., Qwen2.5-Math-7B) to the registry
would likely be the natural fix.

Table~\ref{tab:efficiency} shows \grade{} achieving the highest MMLUPro
accuracy at 52GB VRAM and a latency range of 3.1--11.4s.
The range reflects the Depth Gate: easy queries reach only $M_0$
($\sim$3s) while hard ones activate the full sub-agent pool ($\sim$11s).
Puppeteer runs every query at 13s with a fixed 56GB footprint.

\begin{figure*}[t]
  \centering
  \includegraphics[width=\textwidth]{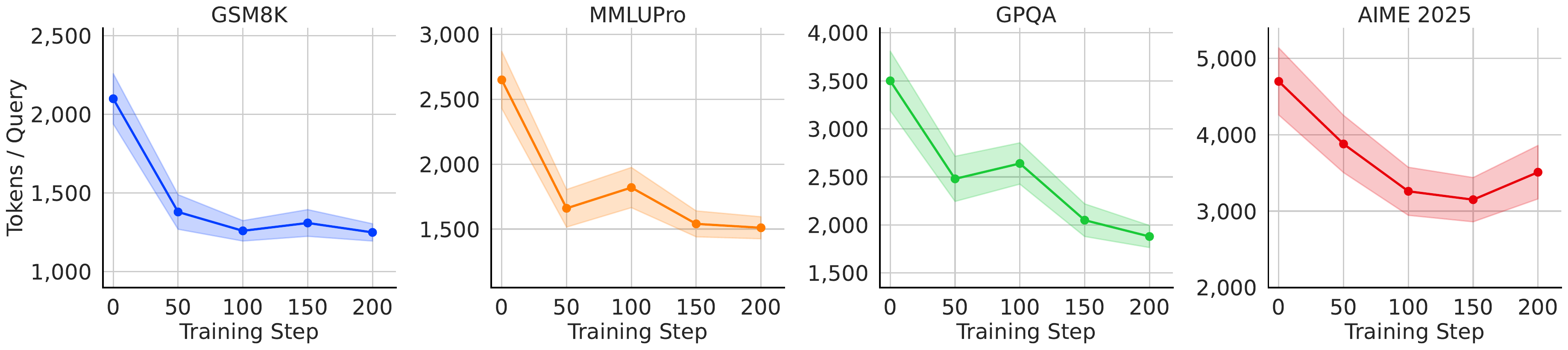}
  \caption{Token usage per query at 50-step training checkpoints ($\pm$1 SD
    shaded), with the settled budget shown as a dashed reference.
    GSM8K converges by step~50; MMLUPro and GPQA show a secondary
    exploration peak near step~100 before settling.
    AIME-2025 does not converge within 200 steps and trends upward
    near the end, confirming that harder families require more training
    to stabilize agent allocation.}
  \label{fig:tokens}
\end{figure*}

\subsection{Component Ablation}
\label{sec:e2}

Table~\ref{tab:ablation_components} presents the results of leave-one-out ablations of different components.

\paragraph{Hierarchy is the most critical component.}
Removing the Depth Gate and manager level drops accuracy by 7.2
points.
Without the Depth Gate, every query pays for the full sub-agent stack,
wasting compute on easy queries and injecting noise from unnecessary
activations.
GPQA suffers most as multi-domain expert questions rely on the
manager layer to group domain-specialist agents before their outputs
are combined.

\paragraph{Cross-Read enables multi-domain synthesis.}
Disabling Cross-Read drops overall accuracy 4.2 points, with the sharpest
decline on GPQA.
The Cross-Read Gate lets agents in different domains share relevant context
before combination; without it, a biology-specialist agent cannot inform a
chemistry-specialist on boundary questions.

\paragraph{Memory and dynamic topology are also important.}
Removing the persistent memory buffer costs 1.7 points.
All benchmarks are single-turn, so memory helps mainly on the
multi-hop compound questions within GPQA.
Switching from Bernoulli routing to a fixed top-$k$ topology costs 
1.1 points, confirming that adaptive allocation provides consistent but
modest gains.

\subsection{Comparison with Different RL Objectives}
\label{sec:e3}

Table~\ref{tab:ablation_rl} compares \cogrpo{} with different RL objectives.

\paragraph{Critic-free vs. Critic-based.}
Both critic-based methods underperform: MAPPO at 60.0 and MAPoRL-obj at
60.6, versus the worst critic-free method (Shared Reward) at 61.6.
Fitting a GAE value function over the combinatorial gate-decision space
converges slowly; the group average is a cheaper and more stable baseline.
MAPoRL-obj closes 0.6 points on MAPPO by adding influence-aware rewards,
but still trails \cogrpo{} by 3.3 points.

\paragraph{Baseline quality within critic-free objectives.}
Among critic-free methods, accuracy tracks baseline quality.
MAGRPO-obj's per-query group mean is stronger than a global mean, but
without std-normalization the gradient magnitude varies with the absolute
reward scale rather than with the relative information content of each
rollout.
The 1.6-point gap between MAGRPO-obj and \cogrpo{} isolates the
contribution of std-normalization and importance-ratio clipping.

On the $k$ sweep, performance peaks at $k{=}3$ and falls at $k{=}5$: a
sixth agent adds noise once the most skilled specialists are already active.
Dynamic routing  slightly outperforms every fixed-$k$ setting by
matching agent count to query difficulty.
\subsection{Hot-Swap Robustness}
\label{sec:e4}

Table~\ref{tab:swap} and Figure~\ref{fig:swap} answer RQ4.
Removing calibration maps is catastrophic -- accuracy falls 18.1 points
immediately (78.2 $\to$ 60.1) and does not recover within 20 evaluation
queries.
Among calibrated swaps, the hardest is API$\to$Local (GPT-4o-mini to
Qwen2.5-7B, 9 queries to recover) - the local model's lower output
confidence shifts the Prune Gate's value estimates.
Same-family swaps (Qwen2.5-7B to Qwen3-7B) recover in 4 queries because
both models share similar pretraining.

The results confirm that \textbf{calibration is necessary for safe
hot-swapping} --- without it, gate thresholds fitted to one model's output
scale silently misfire when a replacement with a different distribution
is swapped in.

\subsection{Cross-Read Frequency}
\label{sec:e5}

The learned Cross-Read Gate settles near $\rho{=}0.5$ in the default
configuration (Figure~\ref{fig:crossread}).
When $\rho{=}1.0$, half the agent pairs produce
noisy or redundant messages, and attending to them hurts performance.
Token usage trajectories differ by task (Figure~\ref{fig:tokens}).
GSM8K converges rapidly --- tokens fall from $\sim$2,100 to the 1,250 budget by
step 50 with only a minor rebound at step 150.
MMLUPro and GPQA both exhibit a secondary peak near step 100 --- a brief
re-exploration phase as the Prune Gate shifts from coarse branch elimination
to fine-grained agent selection --- before settling toward their respective
budgets of 1,500 and 1,860.
AIME-2025 is qualitatively different -- after an initial partial decline, token
usage plateaus around step 100 and then rises through step 200,
reflecting that the gates have not yet learned to reliably prune hard
reasoning chains and may require more training steps to converge.
Table~\ref{tab:agents} shows a clear difficulty-aware allocation: GSM8K
averages 2.1 agents at depth 1.8, often stopping at $M_0$ or the manager
level; AIME-2025 recruits 3.8 agents at depth 2.0, always reaching the
sub-agent pool.
The gates learn to use agent count and hierarchy depth as implicit signals
of query difficulty.

\section{Analysis and Discussion}
\label{sec:analysis}

\paragraph{Gains come from restraint, not added capacity.}
Three independent results point to one mechanism: \grade{} wins by
\emph{withholding} compute, not by adding it. Dynamic assignment beats every
fixed-$k$ setting, and accuracy \emph{falls} from $k{=}3$ ($63.5$) to $k{=}5$
($62.1$) --- a sixth agent injects noise once the strongest specialists are
active (Table~\ref{tab:ablation_rl}). The Cross-Read Gate settles at
$\rho\ {\approx}\ 0.5$, and forcing full communication ($\rho\ {=}\ 1.0$) costs $2.1$
points (Figure~\ref{fig:crossread}). The Depth Gate skips the sub-agent pool
entirely on easy queries. In each case the learned gate beats the \textit{always-on}
setting, and over-provisioning of agents is \emph{harmful} rather than only
wasteful.

\paragraph{Calibration as a safety mechanism.}
When an expert is replaced, the gate thresholds $\tau^A$ and $\tau^P$,
which were fitted to the original model's embedding scale, silently
misfire on the replacement's output distribution.
Per-agent calibration maps $\kappa_a$ absorb this shift using only 64
anchor queries, without touching any other part of the system.

\paragraph{Relation to mixture-of-experts.}
\grade{} differs from sparse MoE
routing~\citep{shazeer2017moe, fedus2021switch} in three ways ---
experts are full pretrained LMs, routing is factored across separate learned gates rather
than a single top-$k$ function, and training uses collaborative RL rather
than supervised prediction with auxiliary losses.

\section{Conclusion}
\label{sec:conclusion}

We presented \grade, a hierarchical multi-agent system that coordinates
heterogeneous expert agents through four jointly trained coordination
gates: Assignment, Depth, Cross-Read, and Prune.
Trained end-to-end with \cogrpo, \grade{} achieves state-of-the-art
results on GSM8K, MMLUPro, and GPQA at approximately 17B average
active parameters, outperforming the strongest baseline
while using just over half the active compute and $\sim$7\% less GPU memory.
On AIME-2025, Puppeteer retains a 1.9-point advantage attributable to a stronger individual model capacity. Our hot-swap analysis establishes per-agent calibration as a necessary condition for safe runtime expert substitution, with uncalibrated swaps causing 18-point accuracy drops and needing more than 20 queries to recover.


\bibliographystyle{IEEEtranN}
\small
\bibliography{refs}

\clearpage
\appendix

\section{Additional Experimental Details}

\subsection{Hyperparameters}
\label{app:hparams}
Table~\ref{tab:hparams} lists the full set of hyperparameters for the main
\grade{} configuration; all experiments use these values unless stated
otherwise.
\begin{table}[!h]
\centering
\caption{Full training hyperparameters for the main experimental
configuration.}
\label{tab:hparams}
\begin{tabular}{lc}
\toprule
\textbf{Hyperparameter} & \textbf{Value} \\
\midrule
Model dimension $d$        & 64 \\
Group size $G$             & 8 \\
Inner epochs (PPO)         & 2 \\
Clip parameter $\eta$      & 0.2 \\
KL weight $\beta$          & 0.02 \\
Cost weight $\lambda_c$          & 0.05 \\
Margin weight $\lambda_m$        & 0.10 \\
Agent-align weight $\lambda_a$   & 0.05 \\
Log-prob clamp                   & $-5.0$ \\
Load-balance $\lambda_{\mathrm{bal}}$ & 0.01 \\
Depth penalty $\lambda_D$  & 0.02 \\
Read sparsity $\lambda_R$  & 0.01 \\
Prune threshold $\tau^P$   & 0.35 \\
Calibration momentum $\alpha$ & 0.05 \\
Anchor query count (swap)  & 64 \\
Adam LR                    & $1 \times 10^{-4}$ \\
Training steps             & 200 \\
Batch size                 & 48 \\
\bottomrule
\end{tabular}

\end{table}

\subsection{Prompt Templates}
\label{app:prompts}
Table~\ref{tab:prompts} specifies the prompt templates used  for different tasks, components, and baselines.

\begin{table*}[t]
\centering
\caption{Prompt templates for \grade{} components and the prompted baselines.
  Slots in braces (e.g.\ \texttt{\{question\}}) are filled per query;
  \texttt{\{format\}} is the benchmark-specific answer-format string in the
  middle block. Frozen sub-agent experts and the orchestrator share the same
  system instruction. }
\label{tab:prompts}
\small
\scalebox{.9}{
\renewcommand{\arraystretch}{1.35}
\begin{tabular}{@{}p{0.3\textwidth} p{0.7\textwidth}@{}}
\toprule
\textbf{Component} & \textbf{Template} \\
\midrule
\multicolumn{2}{@{}l}{\textit{\grade{} components}}\\[2pt]
Sub-agent expert (system) &
{\ttfamily\footnotesize You are a domain expert. Reason step by step, then end with your final answer on a new line as ANSWER: <value>.} \\
Sub-agent expert (user) &
{\ttfamily\footnotesize Question: \{question\}\newline \{format\}} \\
Orchestrator $M_0$ (depth 0) &
{\ttfamily\footnotesize Answer the question directly and concisely. Question: \{question\}\newline \{format\}} \\
\midrule
\multicolumn{2}{@{}l}{\textit{Answer-format slot \texttt{\{format\}} (benchmark-specific)}}\\[2pt]
GSM8K &
{\ttfamily\footnotesize Give only the final numeric answer after ANSWER:, with no units.} \\
MMLUPro &
{\ttfamily\footnotesize Choose one option (A--J). Respond with ANSWER: <letter>.} \\
GPQA &
{\ttfamily\footnotesize Choose one option (A--D). Respond with ANSWER: <letter>.} \\
AIME-2025 &
{\ttfamily\footnotesize The answer is an integer in [0, 999]. Respond with ANSWER: <integer>.} \\
\midrule
\multicolumn{2}{@{}l}{\textit{Prompted baselines}}\\[2pt]
CoT (8-shot) &
{\ttfamily\footnotesize Solve the problem with step-by-step reasoning, following the examples.\newline \{8 worked exemplars\}\newline Question: \{question\}\newline \{format\}} \\
Self-Refine (critique) &
{\ttfamily\footnotesize Review the draft for errors. List concrete problems, or output NO ISSUES if it is correct.\newline Question: \{question\}\newline Draft: \{draft\}} \\
Self-Refine (revise) &
{\ttfamily\footnotesize Revise the solution using the feedback.\newline Question: \{question\}\newline Draft: \{draft\}\newline Feedback: \{feedback\}\newline \{format\}} \\
\bottomrule
\end{tabular}}

\end{table*}

\subsection{Agent Pool Specification}
\label{app:agents}
Table~\ref{tab:agents_spec} specifies the sub-agents, their 
models, and the resulting memory footprint. Agents that resolve to the same
expert share a single in-memory copy.
\begin{table*}[t]
\centering
\caption{Agent pool specification. ``Shared'' experts load a single
  model to memory regardless of how many agents reference them. VRAM is measured on a NVIDIA H200 141GB GPU.}
\label{tab:agents_spec}
\scalebox{0.90}{
\begin{tabular}{l l l c}
\toprule
\textbf{Agent(s)} & \textbf{Backend} & \textbf{Provider} & \textbf{Params} \\
\midrule
$M_0$ (orchestrator)     & Qwen2.5-0.5B-Instruct & HuggingFace & 0.5B \\
Manager 1, 2             & MLP grouping module   & ---         & trainable head \\
A1, A2, A3 (shared)      & Qwen2.5-7B-Instruct   & HuggingFace & 7B (×1 copy) \\
B1, B2 (shared)          & Phi-3-mini-4K-Instruct & HuggingFace & 3.8B (×1 copy) \\
B3                       & Llama-3.2-3B-Instruct  & HuggingFace & 3B \\
\midrule
\multicolumn{3}{l}{Peak VRAM } & $\sim$52 GB \\
\multicolumn{3}{l}{Avg.\ active / query} & $\sim$17B \\
\bottomrule
\end{tabular}}

\end{table*}


\section{Additional Figures and Tables}

Table~\ref{tab:ablation_components} reports the full
leave-one-out component ablation, Table~\ref{tab:agents} reports the mean agent count
and hierarchy depth per task, and Table~\ref{tab:algo_compare} reports a
feature-level comparison of \cogrpo{} with related policy-optimization
algorithms. Table~\ref{tab:significance} reports the significance tests of
\grade{} against the strongest baseline.

\begin{table*}[!h]
\centering
\caption{Mean agents recruited and mean hierarchy depth per task.
  Harder benchmarks reliably recruit more agents and reach the deepest
  hierarchy level, confirming that the gates learn difficulty-aware
  allocation.}
\label{tab:agents}
\scalebox{0.9}{
\begin{tabular}{l ccc}
\toprule
\textbf{Task} & \textbf{\#Agents} & \textbf{Std} & \textbf{Mean depth} \\
\midrule
GSM8K     & 2.1 & 0.3 & 1.8 \\
MMLUPro   & 2.3 & 0.3 & 1.9 \\
GPQA      & 3.1 & 0.4 & 2.0 \\
AIME-2025 & 3.8 & 0.4 & 2.0 \\
\bottomrule
\end{tabular}}

\end{table*}

\begin{table*}[!h]
\centering
\caption{Component ablation (leave-one-out). Each row removes one component
  from the full system and retrains from scratch. Hierarchy removal has the
  largest overall impact ($-7.2$ points), followed by Cross-Read ($-4.2$
  points) and Cross-Attention ($-3.1$ points).}
\label{tab:ablation_components}
\scalebox{0.88}{
\begin{tabular}{l cccc c}
\toprule
\textbf{Variant} & \textbf{GSM8K} & \textbf{MMLUPro} &
  \textbf{GPQA} & \textbf{AIME-2025} & \textbf{Overall} \\
\midrule
Full Setup        & \pmv{94.8}{0.3} & \pmv{78.2}{0.6} & \pmv{57.4}{1.8} & \pmv{25.3}{3.6} & \pmv{63.9}{1.6} \\
$-$ Hierarchy         & \pmv{92.5}{0.5} & \pmv{68.4}{0.8} & \pmv{46.8}{2.1} & \pmv{19.2}{3.8} & \pmv{56.7}{1.8} \\
$-$ Memory            & \pmv{94.1}{0.4} & \pmv{75.9}{0.7} & \pmv{54.8}{1.9} & \pmv{23.9}{3.7} & \pmv{62.2}{1.7} \\
$-$ Cross-Read        & \pmv{93.7}{0.4} & \pmv{72.8}{0.7} & \pmv{50.3}{1.9} & \pmv{22.1}{3.7} & \pmv{59.7}{1.7} \\
$-$ Cross-Attention        & \pmv{94.0}{0.4} & \pmv{74.3}{0.7} & \pmv{52.1}{1.9} & \pmv{22.8}{3.6} & \pmv{60.8}{1.7} \\
$-$ Dynamic Topology  & \pmv{94.4}{0.3} & \pmv{76.7}{0.6} & \pmv{55.9}{1.8} & \pmv{24.1}{3.6} & \pmv{62.8}{1.6} \\
\bottomrule
\end{tabular}}

\end{table*}

\begin{table*}[!h]
\centering
\caption{Algorithmic comparison. \textbf{StdNorm}: whether
  within-group standard deviation is used to normalize advantages.
  \textbf{Clipping}: importance-ratio clipping.
  \textbf{Multi-agent}: flat = same-level homogeneous pool; hier.\ =
 heterogeneous hierarchy.
  \cogrpo{} is the only critic-free, standard-normalized, clipped objective
  designed for hierarchical multi-agent coordination.}
\label{tab:algo_compare}
\scalebox{0.82}{
\begin{tabular}{l c c c c c}
\toprule
\textbf{Algorithm} & \textbf{Critic} & \textbf{StdNorm} &
  \textbf{Clipping} & \textbf{KL} & \textbf{Multi-agent} \\
\midrule
REINFORCE               & \xmark & \xmark & \xmark & \xmark & \xmark \\
PPO~\citep{schulman2017ppo}   & \cmark & \cmark & \cmark & opt.   & \xmark \\
GRPO~\citep{sheng2024grpo}    & \xmark & \cmark & \cmark & \cmark & \xmark \\
DeepSeek-R1~\citep{guo2025deepseekr1} & \xmark & \cmark & \cmark & \cmark & \xmark \\
MAGRPO~\citep{liu2025magrpo}  & \xmark & \xmark & \xmark & \xmark & flat \\
MAPoRL~\citep{park2025maporl} & \cmark & \cmark & \cmark & \cmark & flat \\
\midrule
\textbf{\cogrpo{} (Ours)} & \xmark & \cmark & \cmark & \cmark & hier. \\
\bottomrule
\end{tabular}}

\end{table*}

\begin{table*}[t]
\centering
\caption{Statistical significance of \grade{} compared with Puppeteer
(the strongest baseline) on each benchmark. A pooled two-sample
(equal-variance) $t$-test was performed using the three-seed summary
statistics (mean and standard deviation). $\Delta$ denotes the mean
accuracy difference (\grade{} $-$ Puppeteer), and
$p$ is the two-sided $p$-value from the per-benchmark test (each benchmark
is a single \grade{}-vs-Puppeteer comparison, multiple-comparison
correction not applied). Significance levels are
$^{*}p<0.05$, $^{**}p<0.01$, $^{***}p\le0.001$; ns denotes not significant.}
\label{tab:significance}
\scalebox{0.92}{
\begin{tabular}{lccccc}
\toprule
\textbf{Benchmark} & $\Delta$ & $t$ & $\mathit{df}$ & $p$ & \textbf{Sig.} \\
\midrule
GSM8K & $+1.1$ & $4.49$ & $4.0$ & $0.011$ & * \\
MMLUPro & $+4.8$ & $9.80$ & $4.0$ & $0.001$ & *** \\
GPQA & $+3.3$ & $2.18$ & $4.0$ & $0.095$ & ns \\
AIME-2025 & $-1.9$ & $-0.63$ & $4.0$ & $0.899$ & ns \\
\bottomrule
\end{tabular}}

\end{table*}







  

\end{document}

%% file: math_commands.tex
\usepackage{amsmath,amsfonts,bm}

\def\eqref#1{equation~\ref{#1}}

\def\1{\bm{1}}

\DeclareMathAlphabet{\mathsfit}{\encodingdefault}{\sfdefault}{m}{sl}
\SetMathAlphabet{\mathsfit}{bold}{\encodingdefault}{\sfdefault}{bx}{n}



%% file: refs.bib
@article{
wei2022emergent,
title={{Emergent Abilities of Large Language Models}},
author={Jason Wei and Yi Tay and Rishi Bommasani and Colin Raffel and Barret Zoph and Sebastian Borgeaud and Dani Yogatama and Maarten Bosma and Denny Zhou and Donald Metzler and Ed H. Chi and Tatsunori Hashimoto and Oriol Vinyals and Percy Liang and Jeff Dean and William Fedus},
journal={Transactions on Machine Learning Research},
issn={2835-8856},
year={2022},
url={https://openreview.net/forum?id=yzkSU5zdwD},
note={Survey Certification}
}

@inproceedings{wei2022cot,
 author = {Wei, Jason and Wang, Xuezhi and Schuurmans, Dale and Bosma, Maarten and ichter, brian and Xia, Fei and Chi, Ed and Le, Quoc V and Zhou, Denny},
 booktitle = {Advances in Neural Information Processing Systems},
 editor = {S. Koyejo and S. Mohamed and A. Agarwal and D. Belgrave and K. Cho and A. Oh},
 pages = {24824--24837},
 title = {{Chain-of-Thought Prompting Elicits Reasoning in Large Language Models}},
 url = {https://proceedings.neurips.cc/paper_files/paper/2022/file/9d5609613524ecf4f15af0f7b31abca4-Paper-Conference.pdf},
  publisher = {Curran Associates, Inc.},
 volume = {35},
 year = {2022}
}

@inproceedings{madaan2023selfrefine,
 author = {Madaan, Aman and Tandon, Niket and Gupta, Prakhar and Hallinan, Skyler and Gao, Luyu and Wiegreffe, Sarah and Alon, Uri and Dziri, Nouha and Prabhumoye, Shrimai and Yang, Yiming and Gupta, Shashank and Majumder, Bodhisattwa Prasad and Hermann, Katherine and Welleck, Sean and Yazdanbakhsh, Amir and Clark, Peter},
 booktitle = {Advances in Neural Information Processing Systems},
 editor = {A. Oh and T. Naumann and A. Globerson and K. Saenko and M. Hardt and S. Levine},
 pages = {46534--46594},
 title = {{Self-Refine: Iterative Refinement with Self-Feedback}},
 url = {https://proceedings.neurips.cc/paper_files/paper/2023/file/91edff07232fb1b55a505a9e9f6c0ff3-Paper-Conference.pdf},
 volume = {36},
  publisher = {Curran Associates, Inc.},

 year = {2023}
}

@inproceedings{du2023debate,
  title = 	 {{Improving Factuality and Reasoning in Language Models through Multiagent Debate}},
  author =       {Du, Yilun and Li, Shuang and Torralba, Antonio and Tenenbaum, Joshua B. and Mordatch, Igor},
  booktitle = 	 {Proceedings of the 41st International Conference on Machine Learning},
  pages = 	 {11733--11763},
  year = 	 {2024},
  editor = 	 {Salakhutdinov, Ruslan and Kolter, Zico and Heller, Katherine and Weller, Adrian and Oliver, Nuria and Scarlett, Jonathan and Berkenkamp, Felix},
  volume = 	 {235},
  series = 	 {Proceedings of Machine Learning Research},
  month = 	 {21--27 Jul},
  publisher =    {PMLR},
  url = 	 {https://proceedings.mlr.press/v235/du24e.html}
}

@inproceedings{jiang2023llmblender,
    title = "{LLM-Blender: Ensembling Large Language Models with Pairwise Ranking and Generative Fusion}",
    author = "Jiang, Dongfu  and
      Ren, Xiang  and
      Lin, Bill Yuchen",
    editor = "Rogers, Anna  and
      Boyd-Graber, Jordan  and
      Okazaki, Naoaki",
    booktitle = "Proceedings of the 61st Annual Meeting of the Association for Computational Linguistics (Volume 1: Long Papers)",
    month = jul,
    year = "2023",
    address = "Toronto, Canada",
    publisher = "Association for Computational Linguistics",
    url = "https://aclanthology.org/2023.acl-long.792/",
    doi = "10.18653/v1/2023.acl-long.792",
    pages = "14165--14178"
}

@inproceedings{
wang2024moa,
title={{Mixture-of-Agents Enhances Large Language Model Capabilities}},
author={Junlin Wang and Jue Wang and Ben Athiwaratkun and Ce Zhang and James Zou},
booktitle={The Thirteenth International Conference on Learning Representations},
year={2025},
url={https://openreview.net/forum?id=h0ZfDIrj7T}
}

@inproceedings{tang2024evoagent,
    title = "{EvoAgent: Towards Automatic Multi-Agent Generation via Evolutionary Algorithms}",
    author = "Yuan, Siyu  and
      Song, Kaitao  and
      Chen, Jiangjie  and
      Tan, Xu  and
      Li, Dongsheng  and
      Yang, Deqing",
    editor = "Chiruzzo, Luis  and
      Ritter, Alan  and
      Wang, Lu",
    booktitle = "Proceedings of the 2025 Conference of the Nations of the Americas Chapter of the Association for Computational Linguistics: Human Language Technologies (Volume 1: Long Papers)",
    month = apr,
    year = "2025",
    address = "Albuquerque, New Mexico",
    publisher = "Association for Computational Linguistics",
    url = "https://aclanthology.org/2025.naacl-long.315/",
    doi = "10.18653/v1/2025.naacl-long.315",
    pages = "6192--6217",
    ISBN = "979-8-89176-189-6"
}

@inproceedings{dang2025puppeteer,
   author = {Dang, Yufan and Qian, Chen and Luo, Xueheng and Fan, Jingru and Xie, Zihao and Shi, Ruijie and Chen, Weize and Yang, Cheng and Che, Xiaoyin and Tian, Ye and Xiong, Xuantang and Han, Lei and Liu, Zhiyuan and Sun, Maosong},
 booktitle = {Advances in Neural Information Processing Systems},
 editor = {D. Belgrave and C. Zhang and H. Lin and R. Pascanu and P. Koniusz and M. Ghassemi and N. Chen},
 pages = {165025--165059},
 title = {{Multi-Agent Collaboration via Evolving Orchestration}},
  publisher = {Curran Associates, Inc.},
 url = {https://proceedings.neurips.cc/paper_files/paper/2025/file/f1320d2e2842169c6fc89dcbd80e94d0-Paper-Conference.pdf},
 volume = {38},
 year = {2025}
}

@article{sheng2024grpo,
  title     = {{DeepSeekMath}: Pushing the Limits of Mathematical Reasoning in Open Language Models},
      author={Zhihong Shao and Peiyi Wang and Qihao Zhu and Runxin Xu and Junxiao Song and Xiao Bi and Haowei Zhang and Mingchuan Zhang and Y. K. Li and Y. Wu and Daya Guo},
  journal   = {arXiv preprint arXiv:2402.03300v3},
  year      = {2024},
  url       = {https://arxiv.org/abs/2402.03300v3}
}

@article{guo2025deepseekr1,
  title     = {{DeepSeek-R1}: Incentivizing Reasoning Capability in {LLM}s via Reinforcement Learning},
  author={Daya Guo and Dejian Yang and Haowei Zhang and Junxiao Song and Peiyi Wang and Qihao Zhu and Runxin Xu and Ruoyu Zhang and Shirong Ma and Xiao Bi and Xiaokang Zhang and Xingkai Yu and Yu Wu and Z. F. Wu and Zhibin Gou and Zhihong Shao and Zhuoshu Li and Ziyi Gao and Aixin Liu and Bing Xue and Bingxuan Wang and Bochao Wu and Bei Feng and Chengda Lu and Chenggang Zhao and Chengqi Deng and Chenyu Zhang and Chong Ruan and Damai Dai and Deli Chen and Dongjie Ji and Erhang Li and Fangyun Lin and Fucong Dai and Fuli Luo and Guangbo Hao and Guanting Chen and Guowei Li and H. Zhang and Han Bao and Hanwei Xu and Haocheng Wang and Honghui Ding and Huajian Xin and Huazuo Gao and Hui Qu and Hui Li and Jianzhong Guo and Jiashi Li and Jiawei Wang and Jingchang Chen and Jingyang Yuan and Junjie Qiu and Junlong Li and J. L. Cai and Jiaqi Ni and Jian Liang and Jin Chen and Kai Dong and Kai Hu and Kaige Gao and Kang Guan and Kexin Huang and Kuai Yu and Lean Wang and Lecong Zhang and Liang Zhao and Litong Wang and Liyue Zhang and Lei Xu and Leyi Xia and Mingchuan Zhang and Minghua Zhang and Minghui Tang and Meng Li and Miaojun Wang and Mingming Li and Ning Tian and Panpan Huang and Peng Zhang and Qiancheng Wang and Qinyu Chen and Qiushi Du and Ruiqi Ge and Ruisong Zhang and Ruizhe Pan and Runji Wang and R. J. Chen and R. L. Jin and Ruyi Chen and Shanghao Lu and Shangyan Zhou and Shanhuang Chen and Shengfeng Ye and Shiyu Wang and Shuiping Yu and Shunfeng Zhou and Shuting Pan and S. S. Li and Shuang Zhou and Shaoqing Wu and Shengfeng Ye and Tao Yun and Tian Pei and Tianyu Sun and T. Wang and Wangding Zeng and Wanjia Zhao and Wen Liu and Wenfeng Liang and Wenjun Gao and Wenqin Yu and Wentao Zhang and W. L. Xiao and Wei An and Xiaodong Liu and Xiaohan Wang and Xiaokang Chen and Xiaotao Nie and Xin Cheng and Xin Liu and Xin Xie and Xingchao Liu and Xinyu Yang and Xinyuan Li and Xuecheng Su and Xuheng Lin and X. Q. Li and Xiangyue Jin and Xiaojin Shen and Xiaosha Chen and Xiaowen Sun and Xiaoxiang Wang and Xinnan Song and Xinyi Zhou and Xianzu Wang and Xinxia Shan and Y. K. Li and Y. Q. Wang and Y. X. Wei and Yang Zhang and Yanhong Xu and Yao Li and Yao Zhao and Yaofeng Sun and Yaohui Wang and Yi Yu and Yichao Zhang and Yifan Shi and Yiliang Xiong and Ying He and Yishi Piao and Yisong Wang and Yixuan Tan and Yiyang Ma and Yiyuan Liu and Yongqiang Guo and Yuan Ou and Yuduan Wang and Yue Gong and Yuheng Zou and Yujia He and Yunfan Xiong and Yuxiang Luo and Yuxiang You and Yuxuan Liu and Yuyang Zhou and Y. X. Zhu and Yanhong Xu and Yanping Huang and Yaohui Li and Yi Zheng and Yuchen Zhu and Yunxian Ma and Ying Tang and Yukun Zha and Yuting Yan and Z. Z. Ren and Zehui Ren and Zhangli Sha and Zhe Fu and Zhean Xu and Zhenda Xie and Zhengyan Zhang and Zhewen Hao and Zhicheng Ma and Zhigang Yan and Zhiyu Wu and Zihui Gu and Zijia Zhu and Zijun Liu and Zilin Li and Ziwei Xie and Ziyang Song and Zizheng Pan and Zhen Huang and Zhipeng Xu and Zhongyu Zhang and Zhen Zhang},
  journal   = {arXiv preprint arXiv:2501.12948v1},
  year      = {2025},
  url       = {https://arxiv.org/abs/2501.12948v1}
}

@article{raposo2024mod,
  title     = {Mixture of Depths: Dynamically Allocating Compute in Transformer Language Models},
  author    = {Raposo, David and Ritter, Sam and Richards, Blake and Lillicrap, Timothy and Humphreys, Peter Conway and Santoro, Adam},
  journal   = {arXiv preprint arXiv:2404.02258v1},
  year      = {2024},
  url       = {https://arxiv.org/abs/2404.02258v1}
}

@inproceedings{jang2017gumbel,
title={{Categorical Reparameterization with Gumbel-Softmax}},
author={Eric Jang and Shixiang Gu and Ben Poole},
booktitle={International Conference on Learning Representations},
year={2017},
url={https://openreview.net/forum?id=rkE3y85ee}
}

@article{schulman2017ppo,
  title     = {Proximal Policy Optimization Algorithms},
  author    = {Schulman, John and Wolski, Filip and Dhariwal, Prafull and Radford, Alec and Klimov, Oleg},
  journal   = {arXiv preprint arXiv:1707.06347v2},
  year      = {2017},
  url       = {https://arxiv.org/abs/1707.06347v2}
}

@inproceedings{yu2022mappo,
 author = {Yu, Chao and Velu, Akash and Vinitsky, Eugene and Gao, Jiaxuan and Wang, Yu and Bayen, Alexandre and WU, YI},
 booktitle = {Advances in Neural Information Processing Systems},
 editor = {S. Koyejo and S. Mohamed and A. Agarwal and D. Belgrave and K. Cho and A. Oh},
 pages = {24611--24624},
 title = {{The Surprising Effectiveness of PPO in Cooperative Multi-Agent Games}},
  publisher = {Curran Associates, Inc.},
 url = {https://proceedings.neurips.cc/paper_files/paper/2022/file/9c1535a02f0ce079433344e14d910597-Paper-Datasets_and_Benchmarks.pdf},
 volume = {35},
 year = {2022}
}

@inproceedings{ouyang2022rlhf,
 author = {Ouyang, Long and Wu, Jeffrey and Jiang, Xu and Almeida, Diogo and Wainwright, Carroll and Mishkin, Pamela and Zhang, Chong and Agarwal, Sandhini and Slama, Katarina and Ray, Alex and Schulman, John and Hilton, Jacob and Kelton, Fraser and Miller, Luke and Simens, Maddie and Askell, Amanda and Welinder, Peter and Christiano, Paul F and Leike, Jan and Lowe, Ryan},
 booktitle = {Advances in Neural Information Processing Systems},
 editor = {S. Koyejo and S. Mohamed and A. Agarwal and D. Belgrave and K. Cho and A. Oh},
 pages = {27730--27744},
 title = {{Training language models to follow instructions with human feedback}},
  publisher = {Curran Associates, Inc.},
 url = {https://proceedings.neurips.cc/paper_files/paper/2022/file/b1efde53be364a73914f58805a001731-Paper-Conference.pdf},
 volume = {35},
 year = {2022}
}

@inproceedings{
shazeer2017moe,
title={{Outrageously Large Neural Networks: The Sparsely-Gated Mixture-of-Experts Layer}},
author={Noam Shazeer and Azalia Mirhoseini and Krzysztof Maziarz and Andy Davis and Quoc Le and Geoffrey Hinton and Jeff Dean},
booktitle={International Conference on Learning Representations},
year={2017},
url={https://openreview.net/forum?id=B1ckMDqlg}
}

@article{fedus2021switch,
  author  = {William Fedus and Barret Zoph and Noam Shazeer},
title = {{Switch transformers: scaling to trillion parameter models with simple and efficient sparsity}},
year = {2022},
issue_date = {January 2022},
publisher = {JMLR.org},
volume = {23},
issn = {1532-4435},
  journal = {Journal of Machine Learning Research},
month = jan,
articleno = {120},
  number  = {120},

pages   = {1--39},
  url     = {http://jmlr.org/papers/v23/21-0998.html}
}

@article{cobbe2021gsm8k,
  title={Training Verifiers to Solve Math Word Problems},
  author={Cobbe, Karl and Kosaraju, Vineet and Bavarian, Mohammad and Chen, Mark and Jun, Heewoo and Kaiser, Lukasz and Plappert, Matthias and Tworek, Jerry and Hilton, Jacob and Nakano, Reiichiro and Hesse, Christopher and Schulman, John},
  journal={arXiv preprint arXiv:2110.14168v2},
  year={2021},
  url={https://arxiv.org/abs/2110.14168v2}, 
}

@inproceedings{wang2024mmlupro,
 author = {Wang, Yubo and Ma, Xueguang and Zhang, Ge and Ni, Yuansheng and Chandra, Abhranil and Guo, Shiguang and Ren, Weiming and Arulraj, Aaran and He, Xuan and Jiang, Ziyan and Li, Tianle and Ku, Max and Wang, Kai and Zhuang, Alex and Fan, Rongqi and Yue, Xiang and Chen, Wenhu},
 booktitle = {Advances in Neural Information Processing Systems},
 doi = {10.52202/079017-3018},
 editor = {A. Globerson and L. Mackey and D. Belgrave and A. Fan and U. Paquet and J. Tomczak and C. Zhang},
  publisher = {Curran Associates, Inc.},
 pages = {95266--95290},
 title = {{MMLU-Pro: A More Robust and Challenging Multi-Task Language Understanding Benchmark}},
 url = {https://proceedings.neurips.cc/paper_files/paper/2024/file/ad236edc564f3e3156e1b2feafb99a24-Paper-Datasets_and_Benchmarks_Track.pdf},
 volume = {37},
 year = {2024}
}

@inproceedings{rein2023gpqa,
      title={{GPQA: A Graduate-Level Google-Proof Q\&A Benchmark}},
      author={David Rein and Betty Li Hou and Asa Cooper Stickland and Jackson Petty and Richard Yuanzhe Pang and Julien Dirani and Julian Michael and Samuel R. Bowman},
      booktitle={First Conference on Language Modeling},
      year={2024},
      url={https://openreview.net/forum?id=Ti67584b98}
}

@article{meta2024llama3,
  title     = {The {L}lama 3 Herd of Models},
      author={Aaron Grattafiori and Abhimanyu Dubey and Abhinav Jauhri and Abhinav Pandey and Abhishek Kadian and Ahmad Al-Dahle and Aiesha Letman and Akhil Mathur and Alan Schelten and Alex Vaughan and Amy Yang and Angela Fan and Anirudh Goyal and Anthony Hartshorn and Aobo Yang and Archi Mitra and Archie Sravankumar and Artem Korenev and Arthur Hinsvark and Arun Rao and Aston Zhang and Aurelien Rodriguez and Austen Gregerson and Ava Spataru and Baptiste Roziere and Bethany Biron and Binh Tang and Bobbie Chern and Charlotte Caucheteux and Chaya Nayak and Chloe Bi and Chris Marra and Chris McConnell and Christian Keller and Christophe Touret and Chunyang Wu and Corinne Wong and Cristian Canton Ferrer and Cyrus Nikolaidis and Damien Allonsius and Daniel Song and Danielle Pintz and Danny Livshits and Danny Wyatt and David Esiobu and Dhruv Choudhary and Dhruv Mahajan and Diego Garcia-Olano and Diego Perino and Dieuwke Hupkes and Egor Lakomkin and Ehab AlBadawy and Elina Lobanova and Emily Dinan and Eric Michael Smith and Filip Radenovic and Francisco Guzmán and Frank Zhang and Gabriel Synnaeve and Gabrielle Lee and Georgia Lewis Anderson and Govind Thattai and Graeme Nail and Gregoire Mialon and Guan Pang and Guillem Cucurell and Hailey Nguyen and Hannah Korevaar and Hu Xu and Hugo Touvron and Iliyan Zarov and Imanol Arrieta Ibarra and Isabel Kloumann and Ishan Misra and Ivan Evtimov and Jack Zhang and Jade Copet and Jaewon Lee and Jan Geffert and Jana Vranes and Jason Park and Jay Mahadeokar and Jeet Shah and Jelmer van der Linde and Jennifer Billock and Jenny Hong and Jenya Lee and Jeremy Fu and Jianfeng Chi and Jianyu Huang and Jiawen Liu and Jie Wang and Jiecao Yu and Joanna Bitton and Joe Spisak and Jongsoo Park and Joseph Rocca and Joshua Johnstun and Joshua Saxe and Junteng Jia and Kalyan Vasuden Alwala and Karthik Prasad and Kartikeya Upasani and Kate Plawiak and Ke Li and Kenneth Heafield and Kevin Stone and Khalid El-Arini and Krithika Iyer and Kshitiz Malik and Kuenley Chiu and Kunal Bhalla and Kushal Lakhotia and Lauren Rantala-Yeary and Laurens van der Maaten and Lawrence Chen and Liang Tan and Liz Jenkins and Louis Martin and Lovish Madaan and Lubo Malo and Lukas Blecher and Lukas Landzaat and Luke de Oliveira and Madeline Muzzi and Mahesh Pasupuleti and Mannat Singh and Manohar Paluri and Marcin Kardas and Maria Tsimpoukelli and Mathew Oldham and Mathieu Rita and Maya Pavlova and Melanie Kambadur and Mike Lewis and Min Si and Mitesh Kumar Singh and Mona Hassan and Naman Goyal and Narjes Torabi and Nikolay Bashlykov and Nikolay Bogoychev and Niladri Chatterji and Ning Zhang and Olivier Duchenne and Onur Çelebi and Patrick Alrassy and Pengchuan Zhang and Pengwei Li and Petar Vasic and Peter Weng and Prajjwal Bhargava and Pratik Dubal and Praveen Krishnan and Punit Singh Koura and Puxin Xu and Qing He and Qingxiao Dong and Ragavan Srinivasan and Raj Ganapathy and Ramon Calderer and Ricardo Silveira Cabral and Robert Stojnic and Roberta Raileanu and Rohan Maheswari and Rohit Girdhar and Rohit Patel and Romain Sauvestre and Ronnie Polidoro and Roshan Sumbaly and Ross Taylor and Ruan Silva and Rui Hou and Rui Wang and Saghar Hosseini and Sahana Chennabasappa and Sanjay Singh and Sean Bell and Seohyun Sonia Kim and Sergey Edunov and Shaoliang Nie and Sharan Narang and Sharath Raparthy and Sheng Shen and Shengye Wan and Shruti Bhosale and Shun Zhang and Simon Vandenhende and Soumya Batra and Spencer Whitman and Sten Sootla and Stephane Collot and Suchin Gururangan and Sydney Borodinsky and Tamar Herman and Tara Fowler and Tarek Sheasha and Thomas Georgiou and Thomas Scialom and Tobias Speckbacher and Todor Mihaylov and Tong Xiao and Ujjwal Karn and Vedanuj Goswami and Vibhor Gupta and Vignesh Ramanathan and Viktor Kerkez and Vincent Gonguet and Virginie Do and Vish Vogeti and Vítor Albiero and Vladan Petrovic and Weiwei Chu and Wenhan Xiong and Wenyin Fu and Whitney Meers and Xavier Martinet and Xiaodong Wang and Xiaofang Wang and Xiaoqing Ellen Tan and Xide Xia and Xinfeng Xie and Xuchao Jia and Xuewei Wang and Yaelle Goldschlag and Yashesh Gaur and Yasmine Babaei and Yi Wen and Yiwen Song and Yuchen Zhang and Yue Li and Yuning Mao and Zacharie Delpierre Coudert and Zheng Yan and Zhengxing Chen and Zoe Papakipos and Aaditya Singh and Aayushi Srivastava and Abha Jain and Adam Kelsey and Adam Shajnfeld and Adithya Gangidi and Adolfo Victoria and Ahuva Goldstand and Ajay Menon and Ajay Sharma and Alex Boesenberg and Alexei Baevski and Allie Feinstein and Amanda Kallet and Amit Sangani and Amos Teo and Anam Yunus and Andrei Lupu and Andres Alvarado and Andrew Caples and Andrew Gu and Andrew Ho and Andrew Poulton and Andrew Ryan and Ankit Ramchandani and Annie Dong and Annie Franco and Anuj Goyal and Aparajita Saraf and Arkabandhu Chowdhury and Ashley Gabriel and Ashwin Bharambe and Assaf Eisenman and Azadeh Yazdan and Beau James and Ben Maurer and Benjamin Leonhardi and Bernie Huang and Beth Loyd and Beto De Paola and Bhargavi Paranjape and Bing Liu and Bo Wu and Boyu Ni and Braden Hancock and Bram Wasti and Brandon Spence and Brani Stojkovic and Brian Gamido and Britt Montalvo and Carl Parker and Carly Burton and Catalina Mejia and Ce Liu and Changhan Wang and Changkyu Kim and Chao Zhou and Chester Hu and Ching-Hsiang Chu and Chris Cai and Chris Tindal and Christoph Feichtenhofer and Cynthia Gao and Damon Civin and Dana Beaty and Daniel Kreymer and Daniel Li and David Adkins and David Xu and Davide Testuggine and Delia David and Devi Parikh and Diana Liskovich and Didem Foss and Dingkang Wang and Duc Le and Dustin Holland and Edward Dowling and Eissa Jamil and Elaine Montgomery and Eleonora Presani and Emily Hahn and Emily Wood and Eric-Tuan Le and Erik Brinkman and Esteban Arcaute and Evan Dunbar and Evan Smothers and Fei Sun and Felix Kreuk and Feng Tian and Filippos Kokkinos and Firat Ozgenel and Francesco Caggioni and Frank Kanayet and Frank Seide and Gabriela Medina Florez and Gabriella Schwarz and Gada Badeer and Georgia Swee and Gil Halpern and Grant Herman and Grigory Sizov and Guangyi and Zhang and Guna Lakshminarayanan and Hakan Inan and Hamid Shojanazeri and Han Zou and Hannah Wang and Hanwen Zha and Haroun Habeeb and Harrison Rudolph and Helen Suk and Henry Aspegren and Hunter Goldman and Hongyuan Zhan and Ibrahim Damlaj and Igor Molybog and Igor Tufanov and Ilias Leontiadis and Irina-Elena Veliche and Itai Gat and Jake Weissman and James Geboski and James Kohli and Janice Lam and Japhet Asher and Jean-Baptiste Gaya and Jeff Marcus and Jeff Tang and Jennifer Chan and Jenny Zhen and Jeremy Reizenstein and Jeremy Teboul and Jessica Zhong and Jian Jin and Jingyi Yang and Joe Cummings and Jon Carvill and Jon Shepard and Jonathan McPhie and Jonathan Torres and Josh Ginsburg and Junjie Wang and Kai Wu and Kam Hou U and Karan Saxena and Kartikay Khandelwal and Katayoun Zand and Kathy Matosich and Kaushik Veeraraghavan and Kelly Michelena and Keqian Li and Kiran Jagadeesh and Kun Huang and Kunal Chawla and Kyle Huang and Lailin Chen and Lakshya Garg and Lavender A and Leandro Silva and Lee Bell and Lei Zhang and Liangpeng Guo and Licheng Yu and Liron Moshkovich and Luca Wehrstedt and Madian Khabsa and Manav Avalani and Manish Bhatt and Martynas Mankus and Matan Hasson and Matthew Lennie and Matthias Reso and Maxim Groshev and Maxim Naumov and Maya Lathi and Meghan Keneally and Miao Liu and Michael L. Seltzer and Michal Valko and Michelle Restrepo and Mihir Patel and Mik Vyatskov and Mikayel Samvelyan and Mike Clark and Mike Macey and Mike Wang and Miquel Jubert Hermoso and Mo Metanat and Mohammad Rastegari and Munish Bansal and Nandhini Santhanam and Natascha Parks and Natasha White and Navyata Bawa and Nayan Singhal and Nick Egebo and Nicolas Usunier and Nikhil Mehta and Nikolay Pavlovich Laptev and Ning Dong and Norman Cheng and Oleg Chernoguz and Olivia Hart and Omkar Salpekar and Ozlem Kalinli and Parkin Kent and Parth Parekh and Paul Saab and Pavan Balaji and Pedro Rittner and Philip Bontrager and Pierre Roux and Piotr Dollar and Polina Zvyagina and Prashant Ratanchandani and Pritish Yuvraj and Qian Liang and Rachad Alao and Rachel Rodriguez and Rafi Ayub and Raghotham Murthy and Raghu Nayani and Rahul Mitra and Rangaprabhu Parthasarathy and Raymond Li and Rebekkah Hogan and Robin Battey and Rocky Wang and Russ Howes and Ruty Rinott and Sachin Mehta and Sachin Siby and Sai Jayesh Bondu and Samyak Datta and Sara Chugh and Sara Hunt and Sargun Dhillon and Sasha Sidorov and Satadru Pan and Saurabh Mahajan and Saurabh Verma and Seiji Yamamoto and Sharadh Ramaswamy and Shaun Lindsay and Shaun Lindsay and Sheng Feng and Shenghao Lin and Shengxin Cindy Zha and Shishir Patil and Shiva Shankar and Shuqiang Zhang and Shuqiang Zhang and Sinong Wang and Sneha Agarwal and Soji Sajuyigbe and Soumith Chintala and Stephanie Max and Stephen Chen and Steve Kehoe and Steve Satterfield and Sudarshan Govindaprasad and Sumit Gupta and Summer Deng and Sungmin Cho and Sunny Virk and Suraj Subramanian and Sy Choudhury and Sydney Goldman and Tal Remez and Tamar Glaser and Tamara Best and Thilo Koehler and Thomas Robinson and Tianhe Li and Tianjun Zhang and Tim Matthews and Timothy Chou and Tzook Shaked and Varun Vontimitta and Victoria Ajayi and Victoria Montanez and Vijai Mohan and Vinay Satish Kumar and Vishal Mangla and Vlad Ionescu and Vlad Poenaru and Vlad Tiberiu Mihailescu and Vladimir Ivanov and Wei Li and Wenchen Wang and Wenwen Jiang and Wes Bouaziz and Will Constable and Xiaocheng Tang and Xiaojian Wu and Xiaolan Wang and Xilun Wu and Xinbo Gao and Yaniv Kleinman and Yanjun Chen and Ye Hu and Ye Jia and Ye Qi and Yenda Li and Yilin Zhang and Ying Zhang and Yossi Adi and Youngjin Nam and Yu and Wang and Yu Zhao and Yuchen Hao and Yundi Qian and Yunlu Li and Yuzi He and Zach Rait and Zachary DeVito and Zef Rosnbrick and Zhaoduo Wen and Zhenyu Yang and Zhiwei Zhao and Zhiyu Ma},
journal   = {arXiv preprint arXiv:2407.21783v3},
  year      = {2024},
  url       = {https://arxiv.org/abs/2407.21783v3}
}

@techreport{microsoft2024phi3,
author = {Abdin, Marah I and Ade Jacobs, Sam and Awan, Ammar Ahmad and Aneja, Jyoti and Awadallah, Ahmed and Hassan Awadalla, Hany and Bach, Nguyen and Bahree, Amit and Bakhtiari, Arash and Behl, Harkirat and Benhaim, Alon and Bilenko, Misha and Bjorck, Johan and Bubeck, Sébastien and Cai, Martin and Mendes, Caio César Teodoro and Chen, Weizhu and Chaudhary, Vishrav and Chopra, Parul and Giorno, Allie Del and de Rosa, Gustavo and Dixon, Matthew and Eldan, Ronen and Iter, Dan and Goswami, Abhishek and Gunasekar, Suriya and Haider, Emman and Hao, Junheng and Russell J. Hewett and Huynh, Jamie and Javaheripi, Mojan and Jin, Xin and Kauffmann, Piero and Karampatziakis, Nikos and Kim, Dongwoo and Khademi, Mahmoud and Kurilenko, Lev and Lee, James R. and Lee, Yin Tat and Li, Yuanzhi and Liang, Chen and Liu, Weishung and Lin, Xihui (Eric) and Lin, Zeqi and Madan, Piyush and Mitra, Arindam and Modi, Hardik and Nguyen, Anh and Norick, Brandon and Patra, Barun and Perez-Becker, Daniel and Portet, Thomas and Pryzant, Reid and Qin, Heyang and Radmilac, Marko and Rosset, Corby and Roy, Sambudha and Saarikivi, Olli and Saied, Amin and Salim, Adil and Santacroce, Michael and Shah, Shital and Shang, Ning and Sharma, Hiteshi and Song, Xia and Ruwase, Olatunji and Wang, Xin and Ward, Rachel and Wang, Guanhua and Witte, Philipp and Wyatt, Michael and Xu, Can and Xu, Jiahang and Xu, Weijian and Yadav, Sonali and Yang, Fan and Yang, Ziyi and Yu, Donghan and Zhang, Chengruidong and Zhang, Cyril and Zhang, Jianwen and Zhang, Li Lyna and Zhang, Yi and Zhang, Yunan and Zhou, Xiren},
title = {{Phi-3 Technical Report: A Highly Capable Language Model Locally on Your Phone}},
institution = {Microsoft},
year = {2024},
month = {August},
url = {https://www.microsoft.com/en-us/research/publication/phi-3-technical-report-a-highly-capable-language-model-locally-on-your-phone/},
number = {MSR-TR-2024-12},
}

@article{jiang2024mixtral,
  title     = {Mixtral of Experts},
  author={Albert Q. Jiang and Alexandre Sablayrolles and Antoine Roux and Arthur Mensch and Blanche Savary and Chris Bamford and Devendra Singh Chaplot and Diego de las Casas and Emma Bou Hanna and Florian Bressand and Gianna Lengyel and Guillaume Bour and Guillaume Lample and Lélio Renard Lavaud and Lucile Saulnier and Marie-Anne Lachaux and Pierre Stock and Sandeep Subramanian and Sophia Yang and Szymon Antoniak and Teven Le Scao and Théophile Gervet and Thibaut Lavril and Thomas Wang and Timothée Lacroix and William El Sayed},
  journal   = {arXiv preprint arXiv:2401.04088v1},
  year      = {2024},
  url       = {https://arxiv.org/abs/2401.04088v1}
}

@inproceedings{
wu2023autogen,
title={{AutoGen: Enabling Next-Gen {LLM} Applications via Multi-Agent Conversations}},
author={Qingyun Wu and Gagan Bansal and Jieyu Zhang and Yiran Wu and Beibin Li and Erkang Zhu and Li Jiang and Xiaoyun Zhang and Shaokun Zhang and Jiale Liu and Ahmed Hassan Awadallah and Ryen W White and Doug Burger and Chi Wang},
booktitle={First Conference on Language Modeling},
year={2024},
url={https://openreview.net/forum?id=BAakY1hNKS}
}

@inproceedings{
wang2023selfconsistency,
title={{Self-Consistency Improves Chain of Thought Reasoning in Language Models}},
author={Xuezhi Wang and Jason Wei and Dale Schuurmans and Quoc V Le and Ed H. Chi and Sharan Narang and Aakanksha Chowdhery and Denny Zhou},
booktitle={The Eleventh International Conference on Learning Representations },
year={2023},
url={https://openreview.net/forum?id=1PL1NIMMrw}
}

@inproceedings{park2025maporl,
  title = "{MAP}o{RL}: Multi-Agent Post-Co-Training for Collaborative Large Language Models with Reinforcement Learning",
    author = "Park, Chanwoo  and
      Han, Seungju  and
      Guo, Xingzhi  and
      Ozdaglar, Asuman E.  and
      Zhang, Kaiqing  and
      Kim, Joo-Kyung",
    editor = "Che, Wanxiang  and
      Nabende, Joyce  and
      Shutova, Ekaterina  and
      Pilehvar, Mohammad Taher",
    booktitle = "Proceedings of the 63rd Annual Meeting of the Association for Computational Linguistics (Volume 1: Long Papers)",
    month = jul,
    year = "2025",
    address = "Vienna, Austria",
    publisher = "Association for Computational Linguistics",
    url = "https://aclanthology.org/2025.acl-long.1459/",
    doi = "10.18653/v1/2025.acl-long.1459",
    pages = "30215--30248",
    ISBN = "979-8-89176-251-0"
}

@inproceedings{liu2025magrpo,
  title     = {{LLM Collaboration with Multi-Agent Reinforcement Learning}},
  author    = {Liu, Shuo and Liang, Zeyu and Lyu, Xueguang and Amato, Christopher},
  booktitle = {Proceedings of the AAAI Conference on Artificial Intelligence},
  volume    = {40},
  number    = {38},
  pages     = {32150--32158},
  year      = {2026},
  month     = mar,
  doi       = {10.1609/aaai.v40i38.40487},
  url       = {https://doi.org/10.1609/aaai.v40i38.40487},
  publisher = {Association for the Advancement of Artificial Intelligence}
}

@inproceedings{DasGRBPRP19,
  author       = {Abhishek Das and
                  Th{\'{e}}ophile Gervet and
                  Joshua Romoff and
                  Dhruv Batra and
                  Devi Parikh and
                  Mike Rabbat and
                  Joelle Pineau},
  editor       = {Kamalika Chaudhuri and
                  Ruslan Salakhutdinov},
  title        = {{TarMAC: Targeted Multi-Agent Communication}},
  booktitle    = {Proceedings of the 36th International Conference on Machine Learning,
                  {ICML} 2019, 9-15 June 2019, Long Beach, California, {USA}},
  series       = {Proceedings of Machine Learning Research},
  volume       = {97},
  pages        = {1538--1546},
  publisher    = {{PMLR}},
  year         = {2019},
  url          = {http://proceedings.mlr.press/v97/das19a.html},
  bibsource    = {dblp computer science bibliography, https://dblp.org}
}

@inproceedings{
hong2024metagpt,
title={{Meta{GPT}: Meta Programming for A Multi-Agent Collaborative Framework}},
author={Sirui Hong and Mingchen Zhuge and Jonathan Chen and Xiawu Zheng and Yuheng Cheng and Jinlin Wang and Ceyao Zhang and Zili Wang and Steven Ka Shing Yau and Zijuan Lin and Liyang Zhou and Chenyu Ran and Lingfeng Xiao and Chenglin Wu and J{\"u}rgen Schmidhuber},
booktitle={The Twelfth International Conference on Learning Representations},
year={2024},
url={https://openreview.net/forum?id=VtmBAGCN7o}
}

@inproceedings{li2023camel,
 author = {Li, Guohao and Hammoud, Hasan and Itani, Hani and Khizbullin, Dmitrii and Ghanem, Bernard},
 booktitle = {Advances in Neural Information Processing Systems},
 editor = {A. Oh and T. Naumann and A. Globerson and K. Saenko and M. Hardt and S. Levine},
 pages = {51991--52008},
  publisher = {Curran Associates, Inc.},
 title = {{CAMEL: Communicative Agents for "Mind" Exploration of Large Language Model Society}},
 url = {https://proceedings.neurips.cc/paper_files/paper/2023/file/a3621ee907def47c1b952ade25c67698-Paper-Conference.pdf},
 volume = {36},
 year = {2023}
}

@inproceedings{liang2023encouraging,
    title = "{Encouraging Divergent Thinking in Large Language Models through Multi-Agent Debate}",
    author = "Liang, Tian  and
      He, Zhiwei  and
      Jiao, Wenxiang  and
      Wang, Xing  and
      Wang, Yan  and
      Wang, Rui  and
      Yang, Yujiu  and
      Shi, Shuming  and
      Tu, Zhaopeng",
    editor = "Al-Onaizan, Yaser  and
      Bansal, Mohit  and
      Chen, Yun-Nung",
    booktitle = "Proceedings of the 2024 Conference on Empirical Methods in Natural Language Processing",
    month = nov,
    year = "2024",
    address = "Miami, Florida, USA",
    publisher = "Association for Computational Linguistics",
    url = "https://aclanthology.org/2024.emnlp-main.992/",
    doi = "10.18653/v1/2024.emnlp-main.992",
    pages = "17889--17904"
}

@inproceedings{qian2024chatdev,
    title = "{{C}hat{D}ev: Communicative Agents for Software Development}",
    author = "Qian, Chen  and
      Liu, Wei  and
      Liu, Hongzhang  and
      Chen, Nuo  and
      Dang, Yufan  and
      Li, Jiahao  and
      Yang, Cheng  and
      Chen, Weize  and
      Su, Yusheng  and
      Cong, Xin  and
      Xu, Juyuan  and
      Li, Dahai  and
      Liu, Zhiyuan  and
      Sun, Maosong",
    editor = "Ku, Lun-Wei  and
      Martins, Andre  and
      Srikumar, Vivek",
    booktitle = "Proceedings of the 62nd Annual Meeting of the Association for Computational Linguistics (Volume 1: Long Papers)",
    month = aug,
    year = "2024",
    address = "Bangkok, Thailand",
    publisher = "Association for Computational Linguistics",
    url = "https://aclanthology.org/2024.acl-long.810/",
    doi = "10.18653/v1/2024.acl-long.810",
    pages = "15174--15186"
}

@article{chen2023frugalgpt,
title={{Frugal{GPT}: How to Use Large Language Models While Reducing Cost and Improving Performance}},
author={Lingjiao Chen and Matei Zaharia and James Zou},
journal={Transactions on Machine Learning Research},
issn={2835-8856},
year={2024},
url={https://openreview.net/forum?id=cSimKw5p6R}
}

@inproceedings{ong2024routellm,
title={{Route{LLM}: Learning to Route {LLM}s from Preference Data}},
author={Isaac Ong and Amjad Almahairi and Vincent Wu and Wei-Lin Chiang and Tianhao Wu and Joseph E. Gonzalez and M Waleed Kadous and Ion Stoica},
booktitle={The Thirteenth International Conference on Learning Representations},
year={2025},
url={https://openreview.net/forum?id=8sSqNntaMr}
}

@article{graves2016act,
  title={{Adaptive computation time for recurrent neural networks}},
  author={Graves, Alex},
  journal={arXiv preprint arXiv:1603.08983v6},
  year={2016},
  url={https://arxiv.org/abs/1603.08983v6}
}

@inproceedings{dehghani2019universal,
title={{Universal Transformers}},
author={Mostafa Dehghani and Stephan Gouws and Oriol Vinyals and Jakob Uszkoreit and Lukasz Kaiser},
booktitle={International Conference on Learning Representations},
year={2019},
url={https://openreview.net/forum?id=HyzdRiR9Y7},
}

@inproceedings{elbayad2020depth,
title={{Depth-Adaptive Transformer}},
author={Maha Elbayad and Jiatao Gu and Edouard Grave and Michael Auli},
booktitle={International Conference on Learning Representations},
year={2020},
url={https://openreview.net/forum?id=SJg7KhVKPH}
}

@inproceedings{schuster2022confident,
 author = {Schuster, Tal and Fisch, Adam and Gupta, Jai and Dehghani, Mostafa and Bahri, Dara and Tran, Vinh and Tay, Yi and Metzler, Donald},
 booktitle = {Advances in Neural Information Processing Systems},
 editor = {S. Koyejo and S. Mohamed and A. Agarwal and D. Belgrave and K. Cho and A. Oh},
 pages = {17456--17472},
  publisher = {Curran Associates, Inc.},
 title = {{Confident Adaptive Language Modeling}},
 url = {https://proceedings.neurips.cc/paper_files/paper/2022/file/6fac9e316a4ae75ea244ddcef1982c71-Paper-Conference.pdf},
 volume = {35},
 year = {2022}
}

@inproceedings{sukhbaatar2016commnet,
 author = {Sukhbaatar, Sainbayar and szlam, arthur and Fergus, Rob},
 booktitle = {Advances in Neural Information Processing Systems},
 editor = {D. Lee and M. Sugiyama and U. Luxburg and I. Guyon and R. Garnett},
  publisher = {Curran Associates, Inc.},
 pages = {},
 title = {{Learning Multiagent Communication with Backpropagation}},
 url = {https://proceedings.neurips.cc/paper_files/paper/2016/file/55b1927fdafef39c48e5b73b5d61ea60-Paper.pdf},
 volume = {29},
 year = {2016}
}

@inproceedings{foerster2016dial,
 author = {Foerster, Jakob and Assael, Ioannis Alexandros and de Freitas, Nando and Whiteson, Shimon},
 booktitle = {Advances in Neural Information Processing Systems},
 editor = {D. Lee and M. Sugiyama and U. Luxburg and I. Guyon and R. Garnett},
 pages = {},
 title = {{Learning to Communicate with Deep Multi-Agent Reinforcement Learning}},
 url = {https://proceedings.neurips.cc/paper_files/paper/2016/file/c7635bfd99248a2cdef8249ef7bfbef4-Paper.pdf},
 volume = {29},
 year = {2016}
}

@inproceedings{singh2019ic3net,
title={{Individualized Controlled Continuous Communication Model for Multiagent Cooperative and Competitive Tasks}},
author={Amanpreet Singh and Tushar Jain and Sainbayar Sukhbaatar},
booktitle={International Conference on Learning Representations},
year={2019},
url={https://openreview.net/forum?id=rye7knCqK7},
}

@inproceedings{jiang2018atoc,
 author = {Jiang, Jiechuan and Lu, Zongqing},
 booktitle = {Advances in Neural Information Processing Systems},
 editor = {S. Bengio and H. Wallach and H. Larochelle and K. Grauman and N. Cesa-Bianchi and R. Garnett},
 pages = {},
 title = {{Learning Attentional Communication for Multi-Agent Cooperation}},
 url = {https://proceedings.neurips.cc/paper_files/paper/2018/file/6a8018b3a00b69c008601b8becae392b-Paper.pdf},
 volume = {31},
 year = {2018}
}

@inproceedings{guo2017calibration,
  title = 	 {{On Calibration of Modern Neural Networks}},
  author =       {Chuan Guo and Geoff Pleiss and Yu Sun and Kilian Q. Weinberger},
  booktitle = 	 {Proceedings of the 34th International Conference on Machine Learning},
  pages = 	 {1321--1330},
  year = 	 {2017},
  editor = 	 {Precup, Doina and Teh, Yee Whye},
  volume = 	 {70},
  series = 	 {Proceedings of Machine Learning Research},
  month = 	 {06--11 Aug},
  publisher =    {PMLR},
  url = 	 {https://proceedings.mlr.press/v70/guo17a.html}
}

@inproceedings{desai2020calibration,
    title = "{Calibration of Pre-trained Transformers}",
    author = "Desai, Shrey  and
      Durrett, Greg",
    editor = "Webber, Bonnie  and
      Cohn, Trevor  and
      He, Yulan  and
      Liu, Yang",
    booktitle = "Proceedings of the 2020 Conference on Empirical Methods in Natural Language Processing (EMNLP)",
    month = nov,
    year = "2020",
    address = "Online",
    publisher = "Association for Computational Linguistics",
    url = "https://aclanthology.org/2020.emnlp-main.21/",
    doi = "10.18653/v1/2020.emnlp-main.21",
    pages = "295--302"
}

@inproceedings{yao2023react,
title={{ReAct: Synergizing Reasoning and Acting in Language Models}},
author={Shunyu Yao and Jeffrey Zhao and Dian Yu and Nan Du and Izhak Shafran and Karthik R Narasimhan and Yuan Cao},
booktitle={The Eleventh International Conference on Learning Representations },
year={2023},
url={https://openreview.net/forum?id=WE_vluYUL-X}
}

@inproceedings{
dekoninck2026beyond,
title={Beyond Benchmarks: MathArena as an Evaluation Platform for Mathematics with {LLM}s},
author={Jasper Dekoninck and Nikola Jovanovi{\'c} and Tim Gehrunger and K{\'a}ri R{\"o}gnvaldsson and Ivo Petrov and Chenhao Sun and Martin Vechev},
booktitle={3rd AI for Math Workshop: Toward Self-Evolving Scientific Agents at ICML 2026},
year={2026},
url={https://openreview.net/forum?id=DmPE4byHuN}
}
